\documentclass[10pt,twocolumn,letterpaper]{article}

\usepackage[pagenumbers]{cvpr} % To force page numbers, e.g. for an arXiv version
\usepackage{tabularx}   % 支持tabularx环境和X列类型
\usepackage{multirow}   % 支持多行合并\multirow命令
\usepackage{booktabs}   % 提供美观的表格线（toprule等）
\usepackage{amsmath}    % 支持数学符号和\text{}命令
\usepackage{textpos}

\usepackage{makecell}
\usepackage{booktabs}   % For professional-looking tables (\toprule, \midrule, \bottomrule)
\usepackage{graphicx}   % For \rotatebox
\usepackage{array} 

\definecolor{cvprblue}{rgb}{0.21,0.49,0.74}
\usepackage[pagebackref,breaklinks,colorlinks,allcolors=cvprblue]{hyperref}

\def\paperID{43364} % *** Enter the Paper ID here
\def\confName{CVPR}
\def\confYear{2026}

\title{PhysGM: Large Physical Gaussian Model for Feed-Forward 4D Synthesis}

\author{
Chunji Lv$^{1}$\thanks{Equal contribution.} \quad
Zequn Chen$^{2*}$\thanks{Project Leader.} \quad
Donglin Di$^{2}$ \quad
Weinan Zhang$^{3,5}$ \quad
Hao Li$^{2}$ \quad \\
Wei Chen$^{2}$ \quad
Yinjie Lei$^{4}$ \quad
Changsheng Li$^{1}$\thanks{Corresponding author: lcs@bit.edu.cn}
\\
\\
$^{1}$Beijing Institute of Technology \quad
$^{2}$Li Auto \quad
$^{3}$Harbin Institute of Technology\quad \\
$^{4}$Sichuan University\quad
$^{5}$Suzhou Research Institute, Harbin Institute of Technology
\\
{\tt\small 3120250994@bit.edu.cn}\quad
{\tt\small lcs@bit.edu.cn}
}

\begin{document}
% \maketitle
% \thispagestyle{empty} 
% \begin{figure*}[h]
% \setlength{\belowcaptionskip}{-12pt}
% \setlength{\abovecaptionskip}{-3pt}
% \centering
% \includegraphics[width=1.0\textwidth]{intro1} % Reduce the figure size so that it is slightly narrower than the column.
% \caption{Overview of \textbf{PhysGM}. Given a single image, \textbf{PhysGM} performs a single feed-forward pass to directly predict 3D Gaussian Splatting (3DGS) representation and its associated physical properties (e.g., stiffness, mass). This prediction is optimization-free and completes in under one second. The generated parameters then initialize a Material Point Method (MPM) simulator, producing the final, physically plausible 4D animation. }
% \label{fig1}
% \end{figure*}

\twocolumn[{
\maketitle
\begin{center}
\includegraphics[width=0.9\textwidth]{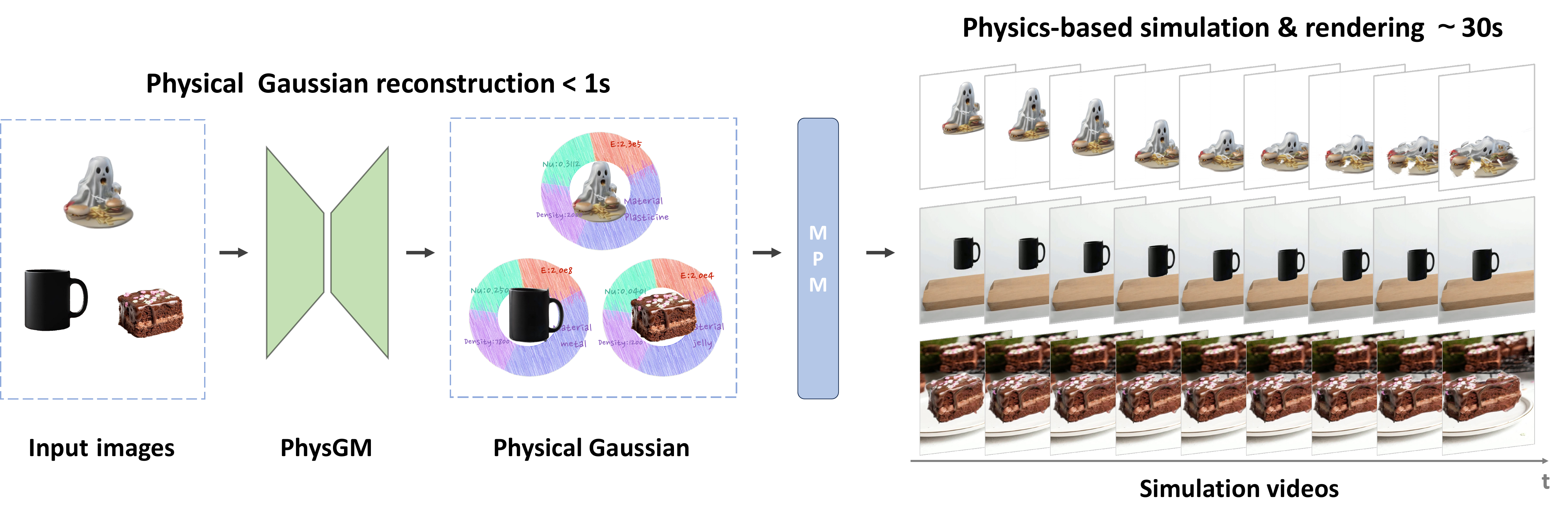}
\end{center}

\captionsetup{type=figure}
\label{fig1}
\captionof{figure}{Overview of \textbf{PhysGM}. Given a single image, \textbf{PhysGM} performs a single feed-forward pass to directly predict 3D Gaussian Splatting (3DGS) representation and its associated physical properties (e.g., stiffness, mass). This prediction is optimization-free and completes in under one second. The generated parameters then initialize a Material Point Method (MPM) simulator, producing the final, physically plausible 4D animation.}

}]

{\let\thefootnote\relax\footnotetext{
\hspace{-2em}
\noindent $^*$Equal contribution.

\noindent $^{\dagger}$Project Leader. 

\noindent $^{\ddagger}$Corresponding author.
}}

\begin{abstract}
Despite advances in physics-based 3D motion synthesis, current methods face key limitations: reliance on pre-reconstructed 3D Gaussian Splatting (3DGS) built from dense multi-view images with time-consuming per-scene optimization; physics integration via either inflexible, hand-specified attributes or unstable, optimization-heavy guidance from video models using Score Distillation Sampling (SDS); and naïve concatenation of prebuilt 3DGS with physics modules, which ignores physical information embedded in appearance and yields suboptimal performance. To address these issues, we propose PhysGM, a feed-forward framework that jointly predicts 3D Gaussian representation and physical properties from a single image, enabling immediate simulation and high-fidelity 4D rendering. Unlike slow appearance-agnostic optimization methods, we first pre-train a physics-aware reconstruction model that directly infers both Gaussian and physical parameters.  We further refine the model with Direct Preference Optimization (DPO), aligning simulations with the physically plausible reference videos and avoiding the high-cost SDS optimization. To address the absence of a supporting dataset for this task, we propose PhysAssets, a dataset of 50K+ 3D assets annotated with physical properties and corresponding reference videos. Experiments show that PhysGM produces high-fidelity 4D simulations from a single image in one minute, achieving a significant speedup over prior work while delivering realistic renderings. Our project page is at:https://hihixiaolv.github.io/PhysGM.github.io/
\end{abstract}   
\vspace{-0.2cm}
\section{Introduction}
\label{sec:intro}
Recent advances in 3D representation, particularly 3D Gaussian Splatting (3DGS)~\cite{kerbl20233d}, have revolutionized novel-view synthesis for static scenes. The next frontier is to imbue these representations with dynamic, physically plausible behavior, unlocking applications in virtual reality, robotics~\cite{lu2024manigaussian}, and autonomous systems~\cite{zhou2024drivinggaussian}. However, creating high-fidelity, physics-based 4D content remains a significant challenge, often demanding computationally expensive, per-scene optimization pipelines that are ill-suited for real-time or large-scale deployment.

Current paradigms for physics-based 4D synthesis are hampered by a fundamental bottleneck: a reliance on slow, iterative optimization. The typical workflow involves first reconstructing a 3DGS model from dense multi-view images, then manually specifying its physical properties (e.g., stiffness, mass) via configuration files~\cite{xie2024physgaussian}, and finally running a physics simulation. This process is not only computationally prohibitive but also lacks scalability and generalization.  While recent works have explored learning physical properties via Score Distillation Sampling (SDS) from video models~\cite{lin2025omniphysgs,zhang2024physdreamer}, which rely on gradient backpropagation through a differentiable physics simulator, incurring heavy per-scene optimization and poor efficiency; hence, they do not resolve the core efficiency bottleneck. Moreover, the common practice of naively concatenating a prebuilt 3DGS with a physics module overlooks physical cues encoded in appearance, causing suboptimal performance.

To face these limitations, we ask a fundamental question: Can we bypass per-scene optimization entirely and instead learn a generative model that produces a complete, physically-grounded 4D simulation in a single, feed-forward pass? This requires reframing the problem from one of slow, iterative reconstruction to one of amortized, feed-forward inference. To this end, we propose \textbf{PhysGM}, a feed-forward framework that enables optimization-free synthesis of dynamic 4D scenes from image inputs. Our key insight is a novel two-stage training paradigm designed to learn a generalizable physical prior and perceptual realism. In the first stage, we pre-train our model on a substantial dataset to jointly predict a 3D Gaussian representation and its corresponding physical properties—this joint optimization not only enhances the accuracy of both geometry and physics predictions but also establishes a well generative prior~\cite{wang2025vggt,dosovitskiy2020image}. Critically, it eliminates the reliance on pre-reconstructed 3DGS that demands multi-view images and time-consuming per-scene optimization, while addressing the flaw of overlooking visually embedded physical information caused by the naive concatenation of separate Gaussian and physics prediction modules. In the second stage, we employ Direct Preference Optimization (DPO)~\cite{rafailov2023direct} to fine-tune the model. By ranking generated simulations against ground-truth videos, we create preference pairs that effectively guide the model towards producing physically plausible and temporally coherent dynamic sequences-unlike SDS-based methods, this preference-driven approach completely eliminates the need for both 
time-consuming per-scene optimization processes and a differentiable physics engine.

Given the lack of datasets pairing 3D assets with physical annotations and reference simulations for this task, we construct and release the PhysAssets Dataset, a substantial benchmark of over 50,000 3D assets, each annotated with its physical material properties and a corresponding physically-plausible simulation video, providing a critical resource for training and evaluating generative 4D models. % A comparison of our approach to prior work is provided in Tab~\ref{tab:intro_table}.

Our main contributions are summarized as follows:
\begin{itemize}
\item We propose PhysGM, the first feed-forward framework capable of generating a physically-grounded 4D Gaussian simulation from an image input in one minute.
\item We construct a novel two-stage training paradigm that combines large-scale supervised pre-training with DPO-based refinement, enabling the model to learn a physical prior and then align it with perceptual quality.
\item We present the PhysAssets Dataset, a new substantial benchmark with 50,000+ 3D assets annotated with physical properties and simulation videos, to spur future research in this challenging domain.
\item Extensive experiments show that PhysGM produces high-fidelity 4D simulations from a single image in one minute, achieving substantial speedups over SDS-based baselines while delivering superior visual quality across metrics.
\end{itemize}

\section{Related Work}
\label{sec:Related Work}
\subsection{4D Content Generation}

Recent advances in 4D content generation—dynamic 3D scenes—primarily follow two paradigms. The first distills knowledge from pre-trained 2D/video diffusion models into a dynamic 3D representation using Score Distillation Sampling (SDS)\cite{poole2022dreamfusion, yin20234dgen, ren2023dreamgaussian4d, ling2024align} or trajectory guidance~\cite{wang2025physctrl}. To ensure temporal coherence, methods employ explicit deformation models\cite{wang2023modelscope} or multi-view video supervision~\cite{zhang20244diffusion, jiang2024animate3d}. However, the iterative nature of SDS is computationally intensive and prone to instability. Mitigation strategies include pre-synthesizing videos for direct supervision~\cite{ren2023dreamgaussian4d, pan2024efficient4d} or training generalizable models to bypass per-scene optimization~\cite{ren2024l4gm,xu20254dgt}.

The second paradigm trains generative models directly on 3D data~\cite{jun2023shap, nichol2022point}. These models have evolved to generate diverse representations, from point clouds~\cite{nichol2022point} to implicit functions (SDFs)\cite{jun2023shap} and explicit 3D Gaussians\cite{tang2024lgm}. Its primary limitation is the scarcity of large-scale 3D datasets.

Despite impressive visual results, both paradigms are fundamentally physics-agnostic. They produce temporally coherent but often physically implausible animations, as they lack an explicit model of real-world dynamics.

\subsection{Feed-forward 3D Gaussian Models}
Parallel to 4D generation, another line of research has focused on accelerating the creation of static 3D scenes. Traditional 3DGS reconstruction~\cite{kerbl20233d} requires slow, per-scene optimization. To address this, feed-forward models have emerged to perform 3D reconstruction from sparse or even single views in a single pass~\cite{szymanowicz2024splatter, tang2024lgm, xu2024grm, zhang2024gs, wu2024direct3d,li2025lirm}. Models like Splatter Image~\cite{szymanowicz2024splatter} and LGM~\cite{tang2024lgm} use U-Net-based architectures to directly regress Gaussian parameters from input images, while LVSM~\cite{jin2024lvsm} outputs only rendered images instead of 3D Gaussians. Other works focus on scene-level generation by aggregating features from multiple images~\cite{charatan2024pixelsplat, chen2024mvsplat, hong2023lrm, li2023instant3d, wang2023pf,ziwen2025long,lin2025dgs}.

However, these feed-forward models have been exclusively developed for static scene generation. They lack any mechanism to represent or predict dynamic, physics-based behavior, limiting their use to non-interactive applications. Our work is the first to embed physical reasoning directly into a feed-forward Gaussian generation framework.

\begin{figure*}[ht]
% \vspace{-8pt}
% \setlength{\belowcaptionskip}{-12pt}
\centering
\includegraphics[width=1.0\textwidth]{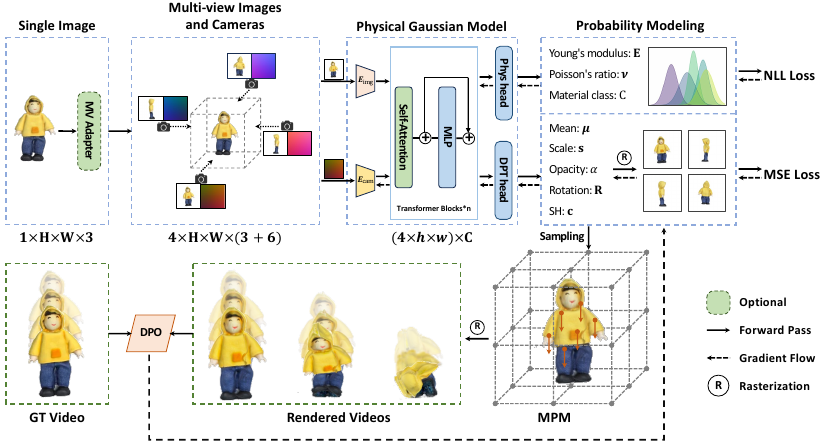} % Reduce the figure size so that it is slightly narrower than the column.
\caption{Pipeline of \textbf{PhysGM}. The model conditions on one or four input views and their corresponding camera parameters, which are processed by a transformer-based model to produce output tokens. These tokens then decoded by two parallel heads: (1) a \textit{DPT Head} predicting the initial 3D Gaussian scene parameters $\boldsymbol{\psi}$, and (2) a \textit{Physics Head} that predicts a distribution over the object's physical properties $\boldsymbol{\theta}$. The sampled parameters ($\boldsymbol{\psi}, \boldsymbol{\theta}$) initialize a Material Point Method (MPM) simulator to generate the final dynamic sequence. The entire architecture is trained in a two-stage paradigm: first, supervised pre-training on ground-truth data establishes a well generative prior. Subsequently, a DPO-based fine-tuning stage uses the ranks against a ground-truth video and  aligns the model with physically  plausible results.}
\label{fig2}
\end{figure*}

\subsection{Physics-Grounded 4D Synthesis}

The integration of physics into 4D generation is a nascent but critical field. The pioneering PhysGaussian~\cite{xie2024physgaussian} first coupled 3DGS with Material Point Method (MPM) simulations based on warp~\cite{macklin2022warp}, but its reliance on manual, per-scene parameter tuning highlighted the need for automation.

Subsequent research has focused on automating this process. A dominant approach uses Score Distillation Sampling (SDS) to distill motion priors from video models, enabling data-driven optimization of material properties~\cite{zhang2024physdreamer, huang2025dreamphysics, liu2024physics3d, tan2024physmotion}. Others leverage Large Language Models (LLMs) to infer parameters from text~\cite{zhao2024efficient, chen2025physgen3d, liu2024physgen, cai2024gic}, though accurately predicting continuous values remains a challenge. More advanced methods learn neural constitutive models~\cite{lin2025omniphysgs}.Concurrently, efforts have been made to construct dedicated physical 3D datasets~\cite{cao2025physx}.

However, existing approaches are fundamentally limited by their reliance on pre-reconstructed models and slow, per-scene optimization. We propose PhysGM to overcome these bottlenecks. Our framework enables rapid, end-to-end 4D synthesis from sparse inputs by learning a physical prior, unifying efficiency with physical realism.

\section{Method}

\label{sec:method}

We present PhysGM, a transformer-based physical reconstruction model that, given the posed RGB images $I=\{I_i \in \mathbb{R}^{H \times W \times 3}\}_{i=1}^n$, the model predicts in one forward pass: (1) a set of 3D Gaussians for geometry and appearance, and (2) a vector of physical properties for physical simulation. We first review the fundamentals of 3D Gaussian Splatting (Sec.~\ref{sec:3dgs}). We then detail our transformer-based architecture that jointly predicts 3DGS and physics parameters (Sec.~\ref{sec:architecture}). Next, we explain how these parameters drive a Material Point Method (MPM) simulation (Sec.~\ref{sec:mpm}) and introduce our DPO-based strategy for refining the model with ground-truth dynamics (Sec.~\ref{sec:dpo_refinement}). Finally, we present PhysAssets, a new dataset created to train and evaluate our model (Sec.~\ref{sec:dataset}).

\subsection{3D Gaussian Splatting}
\label{sec:3dgs} 
The core of 3DGS~\cite{kerbl20233d} is to model the scene as a set of $N$ anisotropic 3D Gaussians. Each Gaussian is parameterized by: {position} $\boldsymbol{\mu} \in \mathbb{R}^3$, defining its center in world space; {covariance} $\boldsymbol{\Sigma} \in \mathbb{R}^{3 \times 3}$ defining its shape and orientation; {opacity} $\alpha \in [0, 1]$ controlling its transparency; {color} represented by Spherical Harmonics (SH) coefficients $\mathbf{c}$ that enable modeling of view-dependent appearance effects. 

The covariance matrix can be computed with a rotation matrix $\mathbf{R_{\text{mat}}}$ and a diagonal scaling matrix $\mathbf{S}$: $\boldsymbol{\Sigma} = \mathbf{R_{\text{mat}}} \mathbf{S} \mathbf{S}^T \mathbf{R_{\text{mat}}}^T $, where $\mathbf{S}=diag(\mathbf{s})$.
During rendering, these 3D Gaussians are projected onto the 2D image plane for a given camera view. The final color $\mathbf{C}$ for a pixel is computed by alpha-blending the $N$ Gaussians that overlap the pixel with its $\alpha_i$ and color $\mathbf{c}_i$ . The objective of all parameters $\psi=\{(\boldsymbol{\mu}_i, \mathbf{q}_i, \mathbf{s}_i, \alpha_i, \mathbf{c}_i)\}_{i=1}^{N}$ is to minimize the difference between the rendered and ground-truth images.

\subsection{Model Architecture}
\label{sec:architecture}
We propose a transformer-based architecture \textbf{PhysGM}, designed to jointly regress 3DGS parameters and physical attributes from posed images. Our framework comprises three main components: multi-modality encoding, a transformer-based backbone, and decoders for 3DGS and physics properties. An overview is shown in Figure~\ref{fig2}.
\vspace{-0.4cm}
\paragraph{Multi-modality tokenization}
% The goal of this module is to encode raw images and camera poses into a sequence of tokens suitable for the transformer backbone. 
% \noindent \textbf{Multi-view Feature Extraction.}
We adopt DINOv3~\cite{siméoni2025dinov3} as image encoder $E_\text{img}$. Each input image $I_i$ is patchified into tokens, which are then projected by a linear layer to produce high-dimensional image features.
% \noindent \textbf{Geometric Embedding.}
To explicitly encode camera geometry, we represent the principal ray of each pixel for view $I_i$ using Plücker ray coordinates $C_{i}$. We process $C_{i}$ by a dense representation encoder~\cite{zhang2025ufmsimplepathunified} $E_{\text{cam}}$ to produce geometry tokens. The resulting camera tokens are then concatenated with corresponding image tokens. The concatenated tokens ${(\textbf{t}_{i})}_{i=1}^{N}$ are formed by concatenating the corresponding image and camera tokens. We further prepend three learnable global tokens ${\mathbf{g}_1,\mathbf{g}_2,\mathbf{g}_3}$ to the sequence. These tokens aggregate global scene information and will be used for physics prediction.
The concatenated tokens $\textbf{t}_{i}$ and input tokens $\mathcal{T}_{\text{in}}$ are thus formed as:
\begin{equation}
\textbf{t}_{i} = \text{concat}(E_{\text{img}}(I_{i}) , E_{\text{cam}}(C_{i})),
\end{equation}
\begin{equation}
\mathcal{T}_{\text{in}} = {(\textbf{t}_{i})}_{i=1}^{N} \cup \mathbf{g}_{k}, k=1,2,3.
\end{equation}

% During inference, when only a single image is available, the pipeline works with three novel views synthesized by MVAdapter~\cite{huang2025mv}. 
During inference with a single input image, we use MVAdapter~\cite{huang2025mv} to synthesize three fixed auxiliary views—rear, left, and right—while the original image serves as the frontal view.

% \noindent \textbf{Global Physics Tokens.}

\vspace{-0.4cm}
\paragraph{Transformer Backbone} 
The complete sequence of input tokens $\mathcal{T}_{\text{in}}$ is fed into our 24-layer transformer backbone to learn the contextualized representations. 
We collect the output tokens from the intermediate layers, which provide a multi-scale representation of the scene, capturing both high-level semantics and low-level details.
\vspace{-0.4cm}
\paragraph{Prediction heads} 
We employ two distinct, specialized heads to map the backbone's output tokens to 3DGS parameters and physical attributes.

To predict the 3DGS representation, we adopt a Dense Prediction Transformer (DPT)~\cite{ranftl2021vision,yang2024depth} head $f_{\text{gs}}$. This head takes the multi-scale features from the backbone and progressively upsamples them through a series of refinement stages. For each input view, it outputs per-pixel maps corresponding to the 3DGS parameters $\psi_i = f_{\text{gs}}(\textbf{t}_i)$. The Gaussians predicted from all views are then aggregated to form the final, coherent 3D scene representation. During pre-training, we optimize the network by jointly minimizing the MSE, alpha and LPIPS~\cite{zhang2018unreasonable} losses between rendered images and the ground-truth views. 

% \textbf{Probabilistic Head for Physics.}
We predict three physical attributes that determine kinetic behavior of the particles: a material class $C$, Young's modulus $E$ (stiffness) and Poisson's ratio $\nu$ (compressibility). We categorize materials into $N_c$ classes, where each class corresponds to a specific constitutive model used in the subsequent MPM simulation. From the three global tokens $\mathbf{g}_k$, we predict these properties using two specialized heads: A classification head $f_{\text{material}}$ to determine the material class; two regression heads $f_{\text{phys}}$ predicting the distribution for the continuous physical properties $E, \nu$ respectively. Specifically, $f_{\text{phys}}$ outputs the mean $\boldsymbol{\mu}_\theta$ and log-variance $\log \boldsymbol{\sigma}_\theta^2$ for these properties:
\begin{equation}
(\boldsymbol{\mu}_\theta, \log \boldsymbol{\sigma}_\theta^2) = f_{\text{phys}}(\mathbf{g}_k).
\end{equation}
This defines a conditional probability distribution over the physical properties, allowing us to model the inherent uncertainty in estimating physics from visual data:
\begin{equation}
P(\boldsymbol{\theta} | I) = \mathcal{N}(\boldsymbol{\theta} | \boldsymbol{\mu}_\theta, \text{diag}(\boldsymbol{\sigma}_\theta^2)).
\label{eq:phys_prob}
\end{equation}
At inference time, we sample from these learned distributions to obtain the scene's parameters: $\boldsymbol{\theta}_{\text{sampled}} \sim P(\boldsymbol{\theta} | I)$. This probabilistic formulation is crucial, as it allows the model to generate diverse physical parameters, enabling the subsequent preference-based refinement with GT videos.

\begin{figure}[ht]
% \vspace{-8pt}
% \setlength{\belowcaptionskip}{-14pt}
\centering
\includegraphics[width=0.4\textwidth]{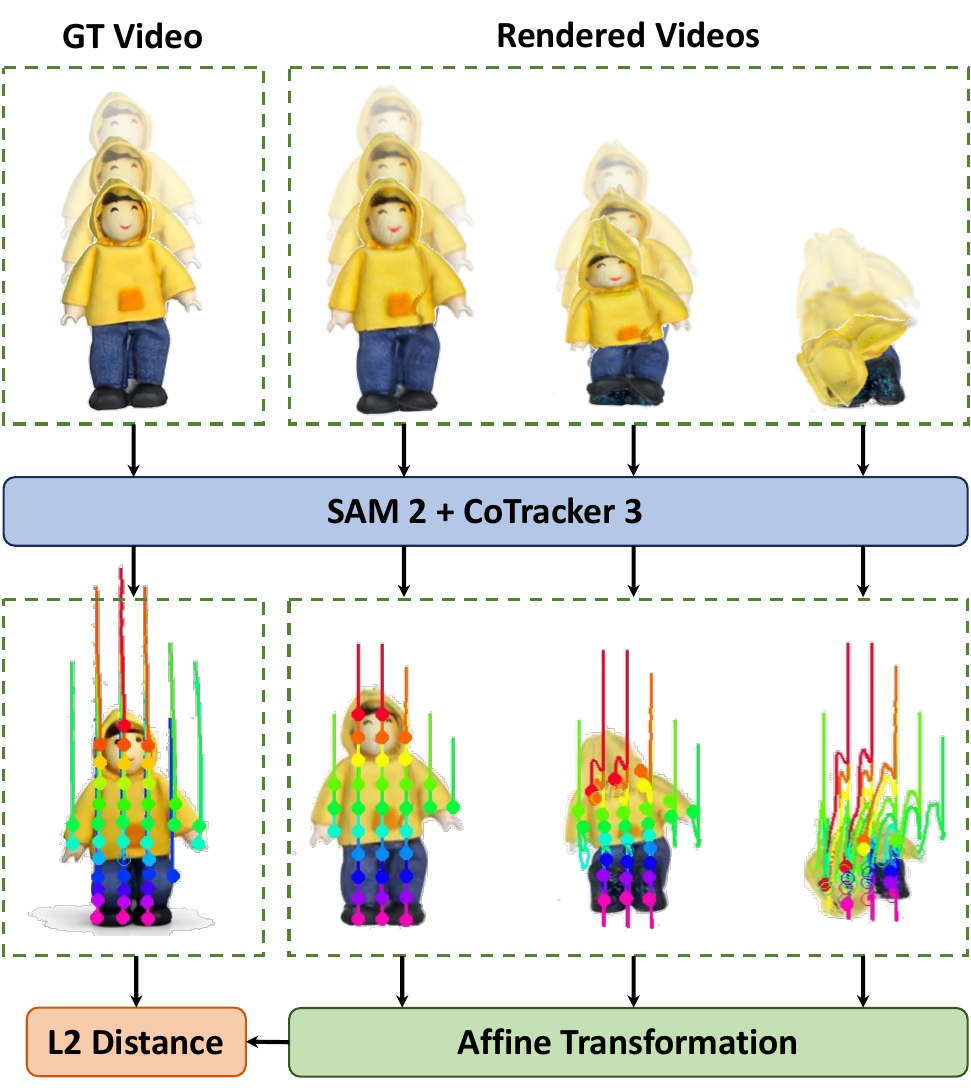} % Reduce the figure size so that it is slightly narrower than the column.
\caption{Preference calculation. We use SAM-2~\cite{ravi2024sam} for segmentation and CoTracker-3 for trajectory extraction across the GT and simulated videos. The extracted point tracks quantify the fidelity of each candidate to the GT, yielding a ranked preference tuple.}
\label{DPO}
\end{figure}

\begin{figure*}[ht]
% \vspace{-8pt}
% \setlength{\belowcaptionskip}{-15pt}
% \setlength{\abovecaptionskip}{-8pt}

\centering
\includegraphics[width=1.0\textwidth]{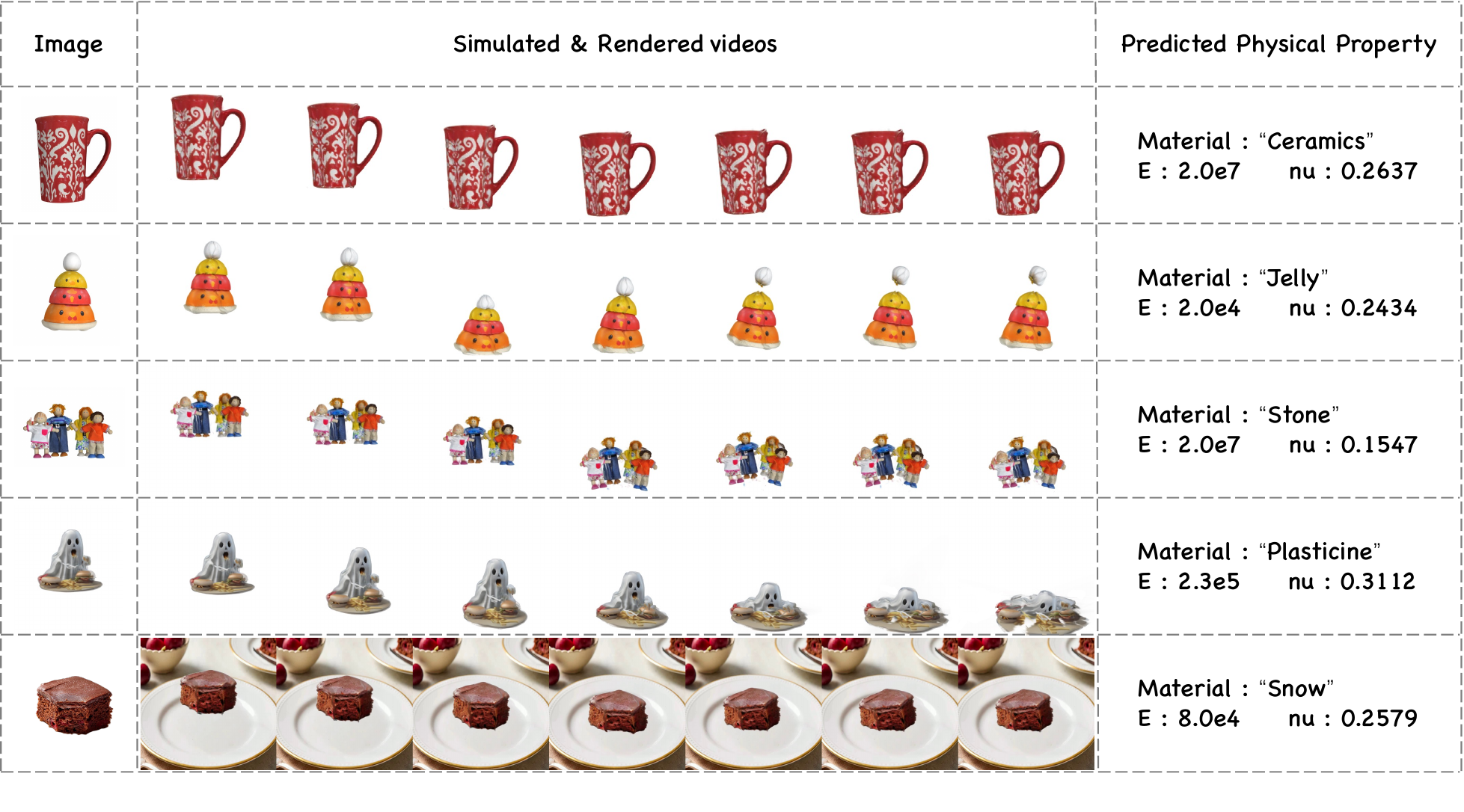} % Reduce the figure size so that it is slightly narrower than the column.
\caption{Qualitative results by \textbf{PhysGM}. For different objects, we show the single input image (left), keyframes from the resulting physically-plausible simulation (middle), and the physical properties predicted by our model (right). Our method generates these high-fidelity 4D sequences in under one minute from a single view, without any per-scene optimization.}
\label{fig3}
% \vspace{-0.2cm}
\end{figure*}

\subsection{Physics-based Dynamics via MPM}
\label{sec:mpm}

To simulate physics-based dynamics, we employ the Material Point Method (MPM)~\cite{jiang2016material,stomakhin2013material}, a hybrid Lagrangian-Eulerian approach that discretizes an object into a set of material points. Each point $p$ carries its own physical state, including mass $m_p$, position $\mathbf{x}_p$, velocity $\mathbf{v}_p$, and the affine velocity matrix $\mathbf{C}_p$. Crucially, it also tracks the deformation gradient $\mathbf{F}_p$, which maps vectors from the material's rest configuration to its current deformed state.

The simulation evolves these states through a two-step process at each time step $\Delta t$. First, in the Particle-to-Grid (P2G) transfer, particle properties are mapped to a background Eulerian grid. The mass $m_i$ and momentum $\mathbf{p}_i$ at each grid node $i$ are computed via weighted summation:
\begin{align}
m_i &= \sum_p m_p N(\mathbf{x}_i - \mathbf{x}_p), \\
\mathbf{p}_i &= \sum_p m_p (\mathbf{v}_p + \mathbf{C}_p (\mathbf{x}_i - \mathbf{x}_p)) N(\mathbf{x}_i - \mathbf{x}_p),
\end{align}
where $N(\cdot)$ is a B-spline interpolation kernel. On the grid, internal forces (derived from $\mathbf{F}_p$ and a constitutive model) and external forces (e.g., gravity) are computed and used to update grid velocities.

Second, in the Grid-to-Particle (G2P) transfer, the updated grid velocity field is interpolated back to update the particle states. The particle's velocity and deformation gradient are updated as follows:
\begin{align}
\mathbf{v}_p^{n+1} &= \sum_i \frac{\mathbf{p}_i^{n+1}}{m_i} N(\mathbf{x}_i - \mathbf{x}_p), \\
\mathbf{F}_p^{n+1} &= \left( \mathbf{I} + \Delta t \sum_i \frac{\mathbf{p}_i^{n+1}}{m_i} \nabla N(\mathbf{x}_i - \mathbf{x}_p)^T \right) \mathbf{F}_p^n.
\end{align}
The particle's final position is $\mathbf{x}_p^{n+1} = \mathbf{x}_p^n + \Delta t \, \mathbf{v}_p^{n+1}$.

Following PhysGaussian~\cite{xie2024physgaussian}, we directly couple this physical simulation with our 3D Gaussian representation. The MPM simulation is driven by the physical properties predicted by our model. We establish a one-to-one correspondence between each material point and a Gaussian primitive. The updated particle position $\mathbf{x}_p$ directly defines the Gaussian's mean $\boldsymbol{\mu}$. More importantly, the deformation gradient $\mathbf{F}_p$ dictates the Gaussian's anisotropic shape and orientation. We perform a polar decomposition on $\mathbf{F}_p$ to factor it into a rotation matrix $\mathbf{R}_p$ and a symmetric positive semi-definite stretch tensor $\mathbf{S}_p$, $\mathbf{F}_p = \mathbf{R}_p \mathbf{S}_p$.
The resulting rotation $\mathbf{R}_p$ and the diagonal elements of the stretch tensor $\mathbf{S}_p$ are then used to define the Gaussian's rotation matrix $\mathbf{R_mat}$ and scaling vector $\mathbf{s}$, respectively. These are used to construct the covariance matrix $\boldsymbol{\Sigma}$, ensuring the rendered geometry mirrors the physically simulated deformation.
\subsection{Preference-based Fine-tuning with DPO}
\label{sec:dpo_refinement}

While supervised pre-training provides a robust physical prior, it may not fully capture the subtle dynamics that lead to high perceptual quality. To bridge this gap, we introduce a fine-tuning stage using Direct Preference Optimization (DPO)~\cite{rafailov2023direct}. This approach enables refining our generative model using feedback from physics simulation and rendering pipeline without requiring differentiability.

We treat the pre-trained model as a fixed reference policy, $\pi_{\text{ref}}$. The model being optimized, $\pi_\omega$, is then refined using a dataset of preference pairs $\mathcal{D} = \{ (\mathbf{z}, \boldsymbol{\phi}_w, \boldsymbol{\phi}_l) \}$. For each scene context $\mathbf{z}$, we perform the following steps to create a preference pair. First, we draw a set of $K$ candidate parameter vectors $\{\boldsymbol{\phi}_1, \dots, \boldsymbol{\phi}_K\}$ from our current policy, where $\boldsymbol{\phi}_k \sim \pi_\omega(\cdot|\mathbf{z})$. Each $\boldsymbol{\phi}$ consists of the predicted physical properties $\theta$ and fixed appearance parameters $\psi$. For each candidate $\boldsymbol{\phi}_k$, we run the MPM simulation and render the resulting 3D Gaussian sequence to produce a short video clip $V_k$.
We compare each generated clip $V_k$ to the ground-truth video $V_{gt}$ using a perceptual distance metric $d(V_k, V_{gt})$.The parameter set $\boldsymbol{\phi}_k$ that yields the lowest perceptual distance is designated the ``winner'' $\boldsymbol{\phi}_w$. Another sample with the highest distance is chosen as the ``loser'' $\boldsymbol{\phi}_l$.

These dynamically generated preference pairs are used to fine-tune $\pi_\omega$ by minimizing the DPO loss. This objective directly increases the likelihood of the model generating the ``winner'' parameters while decreasing the likelihood of the ``loser'' parameters, relative to the reference policy:
\begin{align}
& L_\text{DPO}(\pi_{\omega},\pi_{\text{ref}})=-\mathbb{E}_{(\mathbf{z},\boldsymbol{\phi}_{w},\boldsymbol{\phi}_{l})\sim D}
[\log\sigma(p_{1}-p_{2})], \\
% \end{align}
% where $p_{1}$ and $p_{2}$ are defined as:
% \begin{align}
& p_{1}=\beta \log\frac{\pi_{\omega}(\boldsymbol{\phi}_w|\mathbf{z})}{\pi_{\text{ref}}(\boldsymbol{\phi}_w|\mathbf{z})},\ p_{2}=\beta \log\frac{\pi_{\omega}(\boldsymbol{\phi}_l|\mathbf{z})}{\pi_{\text{ref}}(\boldsymbol{\phi}_l|\mathbf{z})},
\end{align}
where $\beta$ is a temperature parameter controlling the optimization strength. By optimizing this objective, we steer the learned distribution towards regions of the parameter space that produce perceptually and physically superior simulations, without the need for optimization upon complex differentiable MPM and rendering process.

\subsection{PhysAssets Dataset}
\label{sec:dataset}
We introduce PhysAssets, a large dataset of over 50,000 3D objects aggregated from established datasets, including Objaverse~\cite{deitke2023objaverse}, OmniObject3D~\cite{wu2023omniobject3d}, ABO~\cite{collins2022abo} and HSSD~\cite{khanna2023hssd}. To annotate the physical attributes, we developed a pipeline leveraging a large Multimodal Large Language Model (MLLM) Qwen3VL~\cite{qwen3technicalreport}. For each object, we first obtain its corresponding material category directly using a well-designed prompt with multi-view images. Concurrently, we extract the physical property parameters of the object through predefined textual descriptions of Young's modulus and Poisson's ratio. For each object, we generate ground-truth (GT) videos using Framepack~\cite{zhang2025packing}, 
a physics-based simulation framework. The simulation is conditioned on text 
instructions that specify the detailed physical properties of the object, ensuring realistic dynamic behaviors.

To enable DPO training, we automate preference labeling $d(V_k, V_{gt})$ via a comparative analysis pipeline, as shown in Figure~\ref{DPO}. This involves using SAM-2~\cite{kirillov2023segment,ravi2024sam} for instance segmentation and CoTracker-3~\cite{karaev2024cotracker,karaev24cotracker3} for trajectory extraction across the GT and simulated videos. The extracted point tracks then quantify the fidelity of each candidate relative to the GT, yielding a ranked preference tuple. Comprehensive details on the composition of the data set, video generation pipeline, and the preference calculation process are provided in the supplementary materials.

\section{Experiments}
\label{sec:exper}

In this section, we conduct a comprehensive comparison against several baseline methods, evaluating the effectiveness and efficiency of our model. Furthermore, we perform ablation studies on our two-stage training strategy.

\subsection{Experimental settings}
\paragraph{Dataset}
We train our model on our newly created PhysAssets Dataset. For qualitative visualizations, we show results on the test set from our dataset as well as on in-the-wild images to demonstrate generalization.
\vspace{-0.4cm}
\paragraph{Baselines}
We compare our method against two baseline models in physics-based dynamic generation: OmniPhysGS~\cite{lin2025omniphysgs} and DreamerPhysics~\cite{huang2025dreamphysics}. These methods represent the paradigm of per-scene optimization, where physical properties are learned by distilling knowledge from video models using Score Distillation Sampling (SDS). These methods provide strong benchmarks for simulation quality, albeit at a significant computational cost.

\begin{table*}[!t]
% \vspace{-3pt}
% \setlength{\belowcaptionskip}{-8pt}
\centering
\caption{Quantitative comparisons. We evaluate our method and baseline models on 5 different material types. Evaluation is based on the $\text{CLIP}_{sim}$ score (higher is better $\uparrow$) and UPR (higher is better $\uparrow$).}
\small

\renewcommand{\arraystretch}{1.3} 

\begin{tabularx}{\textwidth}{l *{12}{>{\centering\arraybackslash}X}}
\toprule 
\multirow{2}{*}{\raisebox{-0.5\normalbaselineskip}{Method}} & \multicolumn{2}{c}{metal} & \multicolumn{2}{c}{jelly} & \multicolumn{2}{c}{plasticine} & \multicolumn{2}{c}{snow} & \multicolumn{2}{c}{sand} & \multicolumn{2}{c}{average} \\
\cmidrule(lr){2-3} \cmidrule(lr){4-5} \cmidrule(lr){6-7} \cmidrule(lr){8-9} \cmidrule(lr){10-11} \cmidrule(lr){12-13}
& $\text{CLIP}_{sim}$ & UPR & $\text{CLIP}_{sim}$ & UPR & $\text{CLIP}_{sim}$ & UPR & $\text{CLIP}_{sim}$ & UPR & $\text{CLIP}_{sim}$ & UPR & $\text{CLIP}_{sim}$ & UPR \\
\midrule 
OmniPhysGS~\cite{lin2025omniphysgs} 
& 0.2149 & 5\% & 0.2291 & 12\% & 0.2135 & 8\% & 0.1834 & 9\% & 0.2047 & 16\% & 0.2091 & 10\% \\
DreamPhysics~\cite{huang2025dreamphysics}
& 0.2273 & 16\% & 0.2459 & 11\% & 0.2437 & 23\% & 0.2071 & 18\% & 0.2217 & 18\% & 0.2291 & 17.2\% \\
PhysGM (w/o DPO) 
& 0.2698 & 30\% & 0.2700 & 33\% & 0.2547 & 31\% & 0.2541 & 26\% & 0.2980 & 30\% & 0.2693 & 30\% \\
\midrule
PhysGM (w/ DPO) 
& \textbf{0.2732} & \textbf{49\%} & \textbf{0.2774} & \textbf{44\%} & \textbf{0.2691} & \textbf{38\%} & \textbf{0.2548} & \textbf{47\%} & \textbf{0.2997} & \textbf{36\%} & \textbf{0.2748} & \textbf{42.8\%} \\
\bottomrule 
\end{tabularx}

\label{tab:tableTab}
\vspace{-8pt}
\end{table*}

\vspace{-0.4cm}
\paragraph{Simulation Process}
The physical properties predicted by our model directly drive the subsequent Material Point Method (MPM) simulation. The predicted material class determines which constitutive model is employed for the object's dynamics. For instance, a prediction of `rubber' would select a Neo-Hookean constitutive model. The predicted Young's modulus ($E$) and Poisson's ratio ($\nu$) then serve as the specific material parameters for this chosen model. We use a fixed sub-step time of $2 \times 10^{-5}$s and a frame time of $4 \times 10^{-2}$s, generating 50 frames per sequence. Additional simulation parameters, such as boundary conditions, are detailed in the Appendix.

\begin{figure}[htp]
% \setlength{\belowcaptionskip}{-15pt}
% \vspace{-10pt}
\centering
\includegraphics[width=\columnwidth]{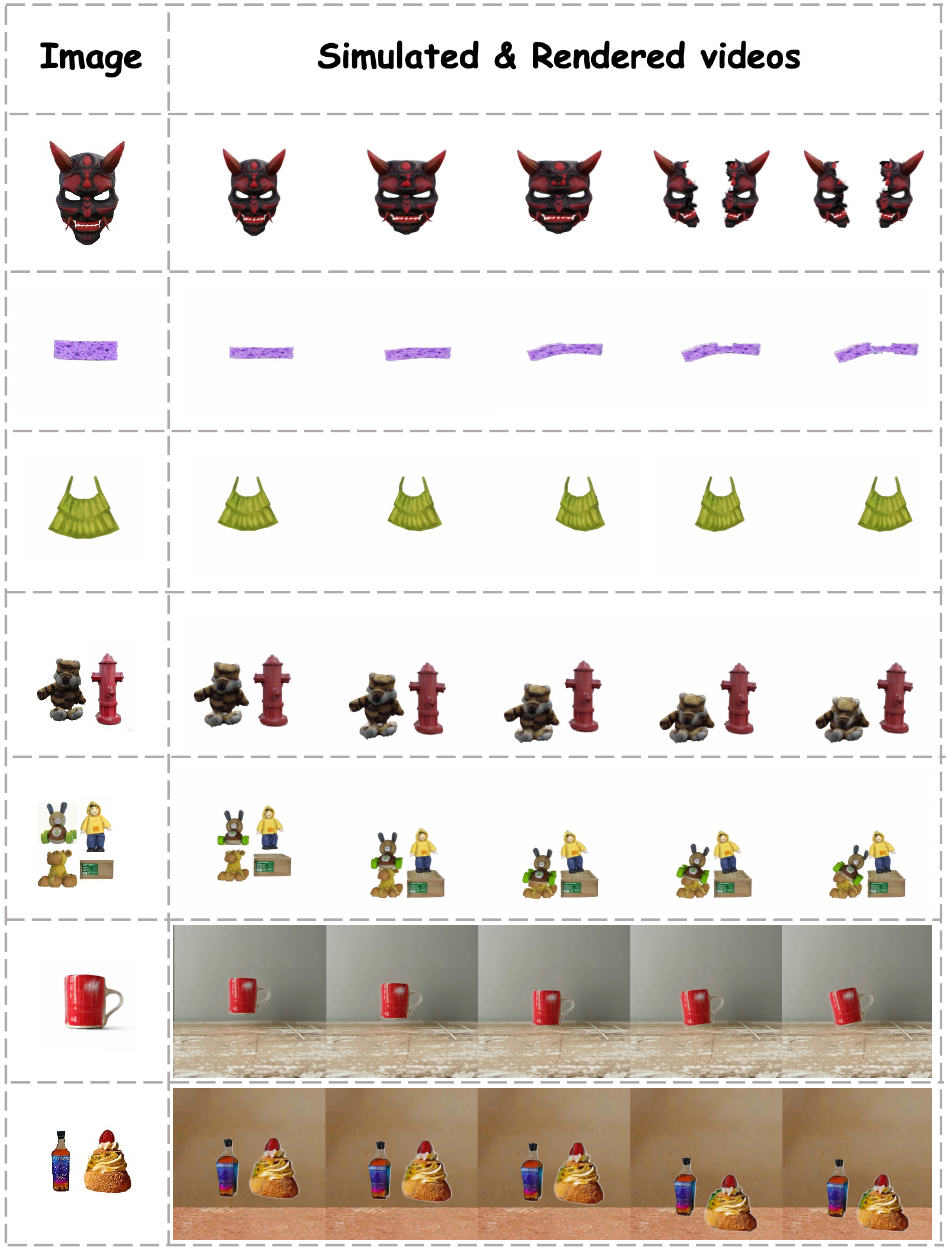} % Reduce the figure size so that it is slightly narrower than the column.
\caption{Other Results by PhysGM. PhysGM can demonstrate robust generalization to diverse physical interactions. It accurately simulates complex deformations like stretching and twisting, handles multi-object dynamics with varied materials, and processes real-world data, highlighting its extensibility to novel scenarios.}

\label{other_results}
% \vspace{-0.4cm}
\end{figure}

\vspace{-0.4cm}
\paragraph{Training Details}
Our model is trained from scratch and leverages FlashAttention v2~\cite{dao2023flashattention2} for efficient computation, with training conducted on 32 NVIDIA A800 GPUs for 3 days and a batch size of 8 per GPU. The training process consists of two stages: in the pre-training phase, we jointly optimize the model for physics property prediction and Gaussian parameter prediction under the supervision of the PhysAssets dataset; In the fine-tuning phase, we employ Direct Preference Optimization (DPO) to further enhance the physical realism and temporal coherence of the generated dynamics, where the model's backbone remains frozen and only the predictive heads responsible for physics properties are fine-tuned.
\vspace{-0.4cm}
\paragraph{Evaluation Metrics}
We conduct a comprehensive evaluation using two complementary metrics: an objective, model-based score and a subjective, human-based assessment.
\textbf{(1)} $\text{CLIP}_{sim}$~\cite{radford2021learning} quantifies the semantic similarity between the rendered visual outcomes and their corresponding textual descriptions of the physical phenomena. \textbf{(2)} User Preference Rate (UPR). We conduct a four-alternative forced-choice user study; UPR is the percentage of trials in which a method is selected as the most realistic or visually appealing. More details are provided in the appendix. \textbf{(3)} We evaluate reconstruction quality with three metrics: PSNR measures pixel-level fidelity via the log ratio of peak signal to MSE (higher is better); SSIM~\cite{wang2004image} assesses structural similarity (0-1, higher is better); and LPIPS~\cite{zhang2018unreasonable} computes learned perceptual distance (lower is better).
\subsection{Results and Analysis}
In this section, we empirically validate the effectiveness of PhysGM. We conduct a comprehensive comparison against baseline methods, and a targeted ablation study to isolate the contributions of our two-stage training paradigm.
\vspace{-0.4cm}
\paragraph{Comparison with baseline methods}
% We begin by showcasing the core capability of optimization-free 4D generation. As illustrated in Figure~\ref{fig3}, our model takes a single input image and, in a single feed-forward pass, generates a complete 3D Gaussian representation along with its corresponding physical properties. This enables a complete physics-based simulation in under one minute. To further demonstrate the robustness and generalization of PhysGM, we present its performance on a variety of diverse and challenging scenarios in Figure~\ref{other_results}.

We demonstrate the core capability of optimization-free 4D generation in Figure~\ref{fig3}. From a single input image, our pipeline generates a 3D Gaussian representation with physical properties via one feed-forward pass, enabling physics-based simulation in under one minute. Figure~\ref{other_results} present its performance on a variety of diverse and challenging scenarios in Figure~\ref{other_results}.

\begin{figure}[ht]

  \centering
  \includegraphics[width=1.0\linewidth]{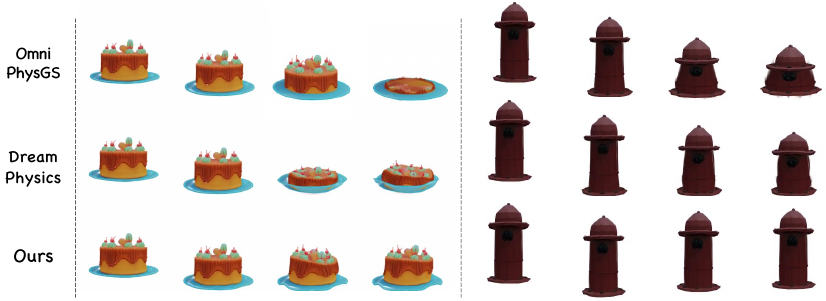}
   \caption{Qualitative comparisons. We selected two distinct physical materials for visual comparison.}
   \label{fig:onecol}

\end{figure}

We evaluated PhysGM against two optimization-based baselines, OmniPhysGS and DreamerPhysics, under identical settings for a fair comparison. Our model used 4-view images as input, while the baselines were initialized with Gaussian splats generated by our model. All physical parameters were kept consistent, except for those each baseline is designed to optimize. Both qualitative results in Figure~\ref{fig:onecol} and quantitative metrics in Table~\ref{tab:tableTab} show that PhysGM outperforms the baselines across a diverse range of materials. This proves our feed-forward approach does not trade quality for speed; by learning a robust physical prior, it surpasses the perceptual realism and physical plausibility of slower, per-scene optimization techniques.

\begin{table}[ht]
\caption{Quantitative comparisons for multi-view synthesis on GSO dataset. We matched the baseline settings by comparing with LGM and GS-LRM, We achieve better results while using only 10\% of the data compared to the GS-LRM.}
  \centering 
    \label{tab:psnr}
  \setlength{\tabcolsep}{4pt} 
  \begin{tabular}{lcccc}  
    \toprule  
    Methods & Res. & PSNR \(\uparrow \) & SSIM \(\uparrow \) & LPIPS \(\downarrow \)  \\  
    \midrule  % 中间细线 
    LGM~\cite{tang2024lgm} & 256 & 21.44 & 0.832 & 0.122 \\
    Our & 256 & \textbf{25.47} & \textbf{0.916} & \textbf{0.071}  \\
    \midrule
    GS-LRM~\cite{zhang2024gs} & 512 & \textbf{30.52} & 0.952 & 0.050  \\
    Ours & 512& 28.95 & \textbf{0.953} & \textbf{0.039}  \\
    \bottomrule  
  \end{tabular}
\vspace{-8pt}
\end{table}

Additionally, we validate MVS effectiveness in Table~\ref{tab:psnr} by 
comparing against Gaussian-based reconstruction methods. Qualitative results 
are in the appendix. Note that our model is trained only at 512 resolution.
% To validate the effectiveness of multi-view stereo (MVS), we conducted experimental evaluations and qualitative visualizations are provided in the appendix.

\vspace{-0.4cm}
\paragraph{Ablation Study}

To validate the effectiveness of our two-stage training strategy, we conduct an ablation focused on the DPO fine-tuning stage. We compare the full PhysGM against a baseline variant (``w/o DPO''), which is trained only with supervised pre-train, also against prior methods.

\begin{figure*}[ht]
% \vspace{-10pt}
% \setlength{\belowcaptionskip}{-15pt}

\centering

\includegraphics[width=1.0\textwidth]{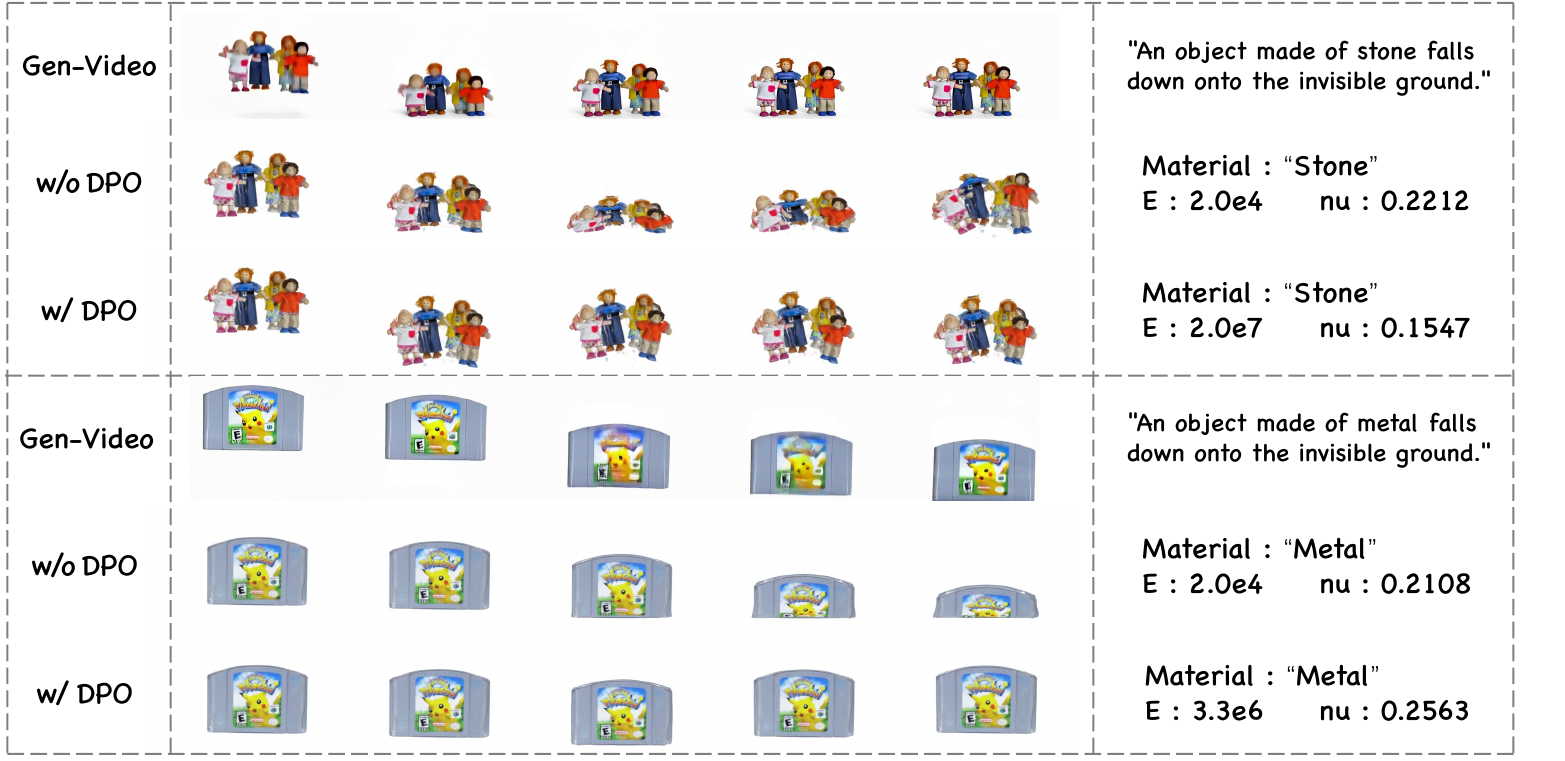} % Reduce the figure size so that it is slightly narrower than the column.
\caption{Ablation results of DPO. The results indicate that after the two-stage DPO training, the model predicts physical attributes with greater accuracy, enabling the generation of 4D videos that exhibit higher physical fidelity.}
\label{fig4}
% \vspace{-5pt}
\end{figure*}

As shown in Table~\ref{tab:with_sds} and Figure~\ref{fig4}, the DPO stage is critical, the full model achieves consistently higher scores across all categories and both metrics. In effect, DPO converts a statistically sound prior into a perceptually superior generator by leveraging feedback from the full, non-differentiable simulation pipeline.

Compared with other methods, PhysGM attains better or comparable visual and physical fidelity at substantially lower optimization cost. Inference requires no per-scene optimization—only a single forward pass followed by MPM rollout—enabling end-to-end generation in under one minute, while competing approaches typically rely on iterative, scene-specific optimization. This yields a markedly better quality–time trade-off.
\begin{table}[ht]
\caption{Comparison with state-of-the-art methods. It can be observed that DPO achieves superior performance in generalization, inference time, and simulation quality.}
  \centering 
    \label{tab:with_sds}
  \setlength{\tabcolsep}{4pt} 
  \begin{tabular}{lcccc}  
    \toprule  
    Methods & Training & Gen. & Infer time & $\text{CLIP}_{sim}$ \\  
    \midrule  % 中间细线
    % PhysGaussian & - & $\times$ & - & 0.2074 \\  
    OmniPhysGS~\cite{lin2025omniphysgs} & SDS & $\times$ & \textgreater 12h & 0.2091 \\
    DreamPhysics~\cite{huang2025dreamphysics} & SDS & $\times$ & \textgreater 0.5h & 0.2291 \\
    \midrule
    Ours & DPO & $\checkmark$ & \textbf{\textless 1min} & \textbf{0.2748} \\
    \bottomrule  
  \end{tabular}
\vspace{-8pt}
\end{table}
\vspace{-1pt}
\section{Discussion}
\vspace{-1pt}
The prohibitive computational cost of the MPM remains the primary obstacle to its use in large-scale, real-time applications, a challenge exacerbated by the lack of viable alternatives for complex physics like fluid and fracture. Critically, the persistent sim-to-real gap—stemming from inherent discrepancies between synthetic training data and physical reality, including simplified constitutive models—hinders robust real-world deployment and limits generalization capabilities. Future work must therefore prioritize two goals: developing more efficient simulation frameworks and bridging the sim-to-real gap.

\vspace{-0.2cm}
\section{Conclusion}
\vspace{-1pt}
We presented \textbf{PhysGM}, a feed-forward framework for rapid, physically grounded 4D synthesis from sparse inputs. Our model is first optimized for 3D Gaussian reconstruction and physical properties prediction, then finetuned with Direct Preference Optimization (DPO) to learn from a non-differentiable simulator, eliminating the need for per-scene optimization. Empirical analysis reveals our approach yields physically realistic simulation and rendering in under a minute. PhysGM's efficiency paves the way for scalable applications in embodied AI, autonomous driving, and interactive virtual reality.
% \clearpage


% \setcounter{page}{1}
% \maketitlesupplementary
\appendix

\newpage
\section*{Appendix}

\begin{table*}[t]
    \centering
    \caption{Comparison with state-of-the-art methods, highlighting PhysGM's unique advantages. Unlike prior work, our method eliminates the need for both pre-optimized 3D Gaussian and pre-defined physical parameters. This allows it to achieve strong generalization while maintaining a significantly shorter inference time ($\textless$ 30s). ``only E'' represents that only Young's modulus is automatically predicted, ``only material'' represents that only material is automatically predicted.}
    \label{tab:sota_comparison}
    \renewcommand{\arraystretch}{1.3}

    \begin{tabular}{@{} l ccccc @{}}
        \toprule
        \textbf{Method} & 
        \makecell{\textbf{No Pre-opt.} \\ \textbf{3D Gaussians}} & 
        \makecell{\textbf{Auto Param} \\ \textbf{Computation}} & 
        \textbf{Generalizable} & 
        \textbf{Without LLM} & 
        \textbf{Inference Time} \\
        \midrule
        PhysGaussian~\cite{xie2024physgaussian} & $\times$ & $\times$ & $\times$ & $\checkmark$ & - \\
        DreamPhysics~\cite{huang2025dreamphysics}      & $\times$     & only E    & $\times$     & $\checkmark$ & \textgreater 0.5h \\
        PhysDreamer~\cite{zhang2024physdreamer}       & $\times$     & only E    & $\times$     & $\checkmark$ & \textgreater 1h   \\
        OmniPhysGS~\cite{lin2025omniphysgs}        & $\times$     & only material    & $\times$     & $\checkmark$ & \textgreater 12h  \\
        DreamGaussian4D~\cite{ren2023dreamgaussian4d}   & $\checkmark$ & $\times$     & $\checkmark$ & $\times$     & 6.5min    \\
        Feature Splatting~\cite{qiu2024feature} & $\times$     & $\times$     & $\times$     & $\checkmark$ & \textgreater 1h   \\
        PhysSplat~\cite{zhao2024efficient}         & $\times$ & $\checkmark$     & $\checkmark$ & $\times$     &  \textless 2min    \\
        \midrule
        \textbf{PhysGM (Ours)} & $\checkmark$ & $\checkmark$ & $\checkmark$ & $\checkmark$ & \textbf{\textless 30s} \\
        
        \bottomrule
    \end{tabular}
\end{table*}

\section{More Details on Implementation}
\subsection{Compare with Other Methods}
Table~\ref{tab:sota_comparison} provides a qualitative comparison between our method, PhysGM, and other state-of-the-art approaches~\cite{xie2024physgaussian,zhang2024physdreamer,huang2025dreamphysics,lin2025omniphysgs,ren2023dreamgaussian4d,qiu2024feature,zhao2024efficient}. We evaluate each method across five critical dimensions: two concerning input requirements (the need for pre-optimized 3D Gaussians or pre-defined physical parameter) and three concerning core capabilities (generalizability, independence from Large Language Models, and inference speed). The comparison highlights that our approach is the only one to operate without these stringent prerequisites. PhysGM simultaneously achieves strong generalization and maintains a very short inference time of under 30 seconds.

\subsection{Simulation Details}
This section elaborates on the key parameters used to configure our Material Point Method (MPM) simulations, as referenced in the main text. The configuration is detailed below, categorized by function.
\vspace{-0.3cm}
\paragraph{MPM Grid Resolution} The simulation domain is discretized into a background grid of 200 * 200 * 200 cells. This grid is fundamental to the MPM algorithm for computing particle interactions and mapping data between particles and the grid.
\vspace{-0.3cm}
\paragraph{Camera Position} For different objects, the camera is initialized at an azimuth of -45 or 135 degrees, an elevation of 0 degrees, and a radius of 1.8 or 1.3 units.
\vspace{-0.3cm}
\paragraph{Camera Motion} The camera is configured to be static during the simulation.
\vspace{-0.3cm}
\paragraph{Other Parameters}Gravity is applied in the falling scene, and force in the corresponding direction is applied in the collision scene.

\subsection{Training and Evaluation Details}
\label{sec:training_details}

\paragraph{Network Architecture.}
We employ DINOv3~\cite{siméoni2025dinov3} (ViT-L/16) pre-trained on LVD-1689M 
as our image encoder, producing 1024-dimensional features. The transformer 
backbone consists of 24 layers with a hidden dimension of 1024 and attention 
head dimension of 64. We use a patch size of 16 and incorporate 3 learnable 
global tokens for physics prediction. For 3D Gaussian representation, we set 
the spherical harmonics degree to 0, with near and far planes at 0.001 and 2.0, 
respectively.
\vspace{-0.3cm}
\paragraph{Training Configuration.}
We train our model on 32 NVIDIA A800 GPUs using a two-stage process for about 3 days in total with a batch size of 8 per GPU. The base learning rate is set to 
$2*10^{-4}$ with AdamW optimizer ($\beta_1=0.9$, $\beta_2=0.95$, weight decay $=$ 0.05). We employ a cosine learning rate schedule with 5K warmup steps and clip gradients to a maximum norm of 10.0. Mixed precision training is enabled using bfloat16 to accelerate computation.
\vspace{-0.3cm}
\paragraph{Data Configuration.}
During training, input images are resized to $512 \times 512$ resolution with square cropping. Each training sample consists of 4 input views and 8 target views, where target views include the input views for consistency. We use 8 workers for data loading with a prefetch factor of 128 to ensure efficient GPU utilization.
\vspace{-0.3cm}
\paragraph{Evaluation Protocol.}
We evaluate on the complete GSO dataset~\cite{downs2022google} containing 
1,009 objects. For each object, we render 32 views with 4 elevation angles 
($0°, 20°, 40°, 60°$) and 8 azimuthal angles. During testing, we sample fixed 4 views as input and evaluate reconstruction quality on 8 randomly 
selected novel views. We report PSNR, SSIM~\cite{wang2004image}, and 
LPIPS~\cite{zhang2018unreasonable} averaged across all test views and objects.

\subsection{User Preference Evaluation}
\label{sec:user_study}

To complement quantitative metrics with human perception assessment, we conduct 
a user study to evaluate the perceptual quality and physical plausibility of 
generated 4D sequences across different methods.
\vspace{-0.3cm}
\paragraph{Study Design.}
We employ a Four-Alternative Forced Choice (4AFC) protocol, where participants are presented with four videos simultaneously showing the same object simulated by different methods: our PhysGM, and three baseline 
methods (PhysGaussian~\cite{xie2024physgaussian}, OmniPhysGS~\cite{lin2025omniphysgs}, 
and DreamGaussian4D~\cite{ren2023dreamgaussian4d}). The videos are displayed 
in randomized positions to eliminate order bias. Participants are instructed 
to select the single video that exhibits the most realistic physical behavior 
and visual quality, considering factors such as motion naturalness, material 
response, temporal coherence, and rendering fidelity.
\vspace{-0.3cm}
\paragraph{Stimuli and Sampling.}
We carefully select 5 representative test scenes spanning diverse object categories and physical scenarios (dropping and stretching). For each scene, we generate 4D sequences using all methods with identical input views and physical interaction setups to ensure fair comparison. 
\vspace{-0.3cm}
\paragraph{Participants and Procedure.}
We recruited 103 participants comprising graduate students and researchers with backgrounds in computer graphics, computer vision, or related fields. Each participant completed a questionnaire containing 5 comparison trials (one per test scene). Before the formal study, participants underwent a training phase with two practice trials to familiarize themselves with the task and interface. Participants could replay videos multiple times before making their selection and were allowed to take breaks between trials. The entire study took approximately 10 minutes per participant.

\paragraph{Data Validation and Filtering.}
To ensure data quality, we implemented several validation mechanisms:
\begin{itemize}[noitemsep,leftmargin=*]
    \item \textbf{Attention checks:} Two control trials with obvious quality 
    differences were inserted to identify inattentive participants.
    \item \textbf{Completion time:} Responses completed too quickly ($<$5 
    seconds per trial) were flagged.
    \item \textbf{Response consistency:} Participants showing random selection 
    patterns were identified via entropy analysis.
\end{itemize}
After applying these criteria, we excluded 3 invalid responses (2.9\% exclusion 
rate) due to failed attention checks or suspiciously short completion times, 
resulting in 100 valid responses for analysis.

\paragraph{User Preference Rate (UPR).}
We define the User Preference Rate as the percentage of participants who selected 
a given method as the most realistic:
\begin{equation}
\text{UPR}_m = \frac{1}{S \cdot N} \sum_{s=1}^{S} \sum_{i=1}^{N} \mathbf{1}[\text{choice}_{s,i} = m] \times 100\%
\end{equation}
where $m$ denotes the method, $S$ is the number of test scenes, $N=100$ is the 
number of valid participants, and $\mathbf{1}[\cdot]$ is the indicator function. 
A higher UPR indicates stronger human preference. Under random chance, each 
method would receive 25\% preference rate in a 4AFC setup.

\section{Material Constitutive Models}
In continuum mechanics and physics-based simulation, a constitutive model (or constitutive equation) is a fundamental mathematical relationship that describes how a material responds to external stimuli. Specifically, it defines the relationship between the internal forces (stress) and the material's deformation (strain). The choice of a constitutive model is critical as it dictates the material's behavior—whether it behaves as a rigid solid, an elastic solid, a fluid, or a hyperelastic material like rubber.Our simulation framework employs different constitutive models based on the predicted material class. This allows us to capture a diverse range of dynamic behaviors.
\subsection{The Neo-Hookean Model} For materials predicted to be ``jelly'' or other soft, rubber-like substances, we employ the Neo-Hookean model. This is a classic hyperelastic model, meaning its stress-response is derived from a strain energy density function. It is ideal for capturing large, nonlinear deformations while remaining computationally efficient, making it a staple in computer graphics and simulation. The model's formulation is based on the statistical mechanics of polymer chains, which accurately describes the behavior of materials like rubber that can stretch significantly without permanent deformation.

The core idea is to split the material's response into two parts: a part that resists changes in shape (deviatoric) and a part that resists changes in volume (volumetric). This allows for a robust simulation of compressible, soft-bodied dynamics. The model defines the Kirchhoff stress ($\tau$), which is a measure of internal force suitable for large-deformation analysis. The Kirchhoff stress $\tau$ for a compressible Neo-Hookean material is given by:
\begin{align}
\tau = \mu * J^{-2/3}*\text{dev}(\mathbf{B})+(\lambda/2)*(J^{2}-1)*I,
\end{align}
where $\tau$ is the Kirchhoff stress tensor. $\mathbf{B}=\mathbf{F}\mathbf{F}^{T}$ is the left Cauchy-Green deformation tensor, where $\mathbf{F}$ is the deformation gradient and $\text{dev}(\mathbf{B})$ is the deviatoric (volume-preserving) part of $\mathbf{B}$. $J=\det(\mathbf{F})$ is the determinant of the deformation gradient, representing the volume change. $\mu$ and $\lambda$ are the Lamé parameters, which characterize the material's stiffness. They are derived from the Young's modulus (E) and Poisson's ratio ($\nu$) predicted by our model.
\subsection{The Fixed Corotational Constitutive Model}
For materials predicted to be ``metal'' or other similarly stiff elastic solids, we employ the Fixed Corotational (FCR) constitutive model. This model is particularly well-suited for scenarios where a material undergoes large rigid-body motions (i.e., translation and rotation) but experiences only small elastic deformations. The core principle of any corotational model is to decouple the object's overall rotation from its internal strain. 
The FCR model begins with the polar decomposition of the deformation gradient $\mathbf{F} = \mathbf{RS}$,
where $\mathbf{R}$ is a pure rotation matrix, and $\mathbf{S}$ is the right stretch tensor, which is symmetric and positive definite.
The model defines a linear relationship between the First Piola-Kirchhoff stress (P) and a measure of strain. The First Piola-Kirchhoff stress is energetically conjugate to the deformation gradient F and is given by:
\begin{align}
\mathbf{P} = 2\mu(\mathbf{F}-\mathbf{R}) + \lambda(J-1)J(\mathbf{F}^{-T}), 
\end{align}
where $\mathbf{P}$ is the First Piola-Kirchhoff stress tensor. For force calculations within our MPM simulation, we use the Kirchhoff stress ($\tau$). The relationship between Kirchhoff stress and the First Piola-Kirchhoff stress is:
$\tau = \mathbf{P}\mathbf{F}^{T}$.

\subsection{The Drucker-Prager Plasticity Model}
For materials exhibiting both frictional and cohesive properties, such as sand, snow, and plasticine, we employ the Drucker-Prager elastoplasticity model. This model is ideal for materials whose strength is dependent on the hydrostatic pressure they are under (e.g., sand becomes stronger when compressed). It defines a yield criterion, which is a surface in stress space that separates elastic (temporary) deformation from plastic (permanent) deformation.
The core of the model is the predictor-corrector algorithm, also known as return mapping: First, the model assumes the material behaves purely elastically during a time step and calculates a ``trial stress''. It then checks if this trial stress lies outside the Drucker-Prager yield surface. If the trial stress is outside the surface (i.e., the material has yielded), the stress is mathematically projected back onto the yield surface. This correction step accounts for the plastic flow and ensures the material's stress state remains physically plausible.
The Drucker-Prager yield criterion defines the boundary between elastic and plastic states. The yield function is given by:
\begin{align}
f(\tau) = ||\text{dev}(\tau)||+\alpha*\text{tr}(\tau)-k \leq 0,
\end{align}
where $\tau$ is the Kirchhoff stress tensor. $\text{dev}(\tau)$ is the deviatoric part of the stress, representing shear. $||\text{dev}(\tau)||$ is the Frobenius norm of the deviatoric stress, measuring the magnitude of the shear stress. $\text{tr}(\tau)$ is the trace of the stress, proportional to the hydrostatic pressure (positive for tension, negative for compression). $\alpha$ is a dimensionless friction parameter, controlling how much the material's strength increases with pressure. $k$ is the cohesion of the material, representing its intrinsic shear strength at zero pressure.

The key insight is that different materials like sand, snow, and plasticine can be simulated with the same underlying model by simply adjusting the cohesion ($k$) and friction ($\alpha$) parameters. For instance: Sand ($k = 0.0$) has negligible cohesion; its strength comes almost entirely from inter-particle friction. Snow ($k = 1000.0$) represents an intermediate case with some cohesion. Plasticine ($k = 5000.0$) has significant cohesion, allowing it to hold its shape even without compressive pressure.

\begin{table}[t]
\centering
\caption{Material distribution in PhysAssets dataset. The 14 primary materials 
account for 97\% of the dataset, while 32 rare materials provide additional 
diversity.}
\label{tab:material_dist}
\small
\begin{tabular}{llrr}
\toprule
\textbf{Rank} & \textbf{Material} & \textbf{Count} & \textbf{Percentage} \\
\midrule
1  & Plastic  & 13,696 & 27.3\% \\
2  & Wood     & 8,443  & 16.8\% \\
3  & Metal    & 7,353  & 14.6\% \\
4  & Fabric   & 7,255  & 14.5\% \\
5  & Ceramic  & 3,023  & 6.0\% \\
6  & Stone    & 2,135  & 4.3\% \\
7  & Paper    & 1,432  & 2.9\% \\
8  & Leather  & 1,132  & 2.3\% \\
9  & Glass    & 955    & 1.9\% \\
10 & Rubber   & 687    & 1.4\% \\
11 & Foam     & 168    & 0.3\% \\
12 & Snow     & 147    & 0.3\% \\
13 & Sand     & 58     & 0.1\% \\
14 & Other (32 materials) & 1,731 & 3.4\% \\
\bottomrule
\end{tabular}
\end{table}

% Google Scanned Objects (GSO)~\cite{downs2022google}
\section{PhysAssets Dataset Statistics}
\label{sec:dataset_stats}

\subsection{Dataset Composition}

PhysAssets comprises a comprehensive collection of 3D assets with annotated 
physical properties. The dataset consists of two main components: a training 
set containing 49,206 objects aggregated from multiple public repositories 
(Objaverse~\cite{deitke2023objaverse}, OmniObject3D~\cite{wu2023omniobject3d}, 
ABO~\cite{collins2022abo}, and HSSD~\cite{khanna2023hssd}), and a held-out 
test set of 1,009 objects from the Google Scanned Objects (GSO) dataset~\cite{downs2022google}, 
totaling 50,215 annotated 3D objects. The primary objective of this effort was to create a comprehensive, diverse, and standardized collection of Physical-based assets annotated with 20+ views rendered images, physical properties, and corresponding guiding videos.

\subsection{Material and Physical Property Distribution}

The dataset exhibits rich material diversity, covering 46 distinct material 
categories. Among these, 14 primary materials constitute the majority of the dataset, while 32 additional rare materials provide coverage for specialized physical scenarios. Table~\ref{tab:material_dist} presents the distribution of the 14 primary materials.

The most represented material is \textbf{Plastic}, with 13,696 samples (27.3\%), 
reflecting its prevalence in manufactured objects. \textbf{Wood} constitutes 
the second largest category with 8,443 samples (16.8\%), followed by 
\textbf{Metal} with 7,353 samples (14.6\%) and \textbf{Fabric} with 7,255 
samples (14.5\%). \textbf{Ceramic} objects account for 3,023 samples (6.0\%). 
Medium-frequency materials include \textbf{Stone} (2,135 samples, 4.3\%), 
\textbf{Paper} (1,432 samples, 2.9\%), \textbf{Leather} (1,132 samples, 2.3\%), 
\textbf{Glass} (955 samples, 1.9\%), and \textbf{Rubber} (687 samples, 1.4\%). 
Low-frequency but physically interesting materials comprise \textbf{Foam} 
(168 samples, 0.3\%), \textbf{Snow} (147 samples, 0.3\%), and \textbf{Sand} 
(58 samples, 0.1\%). The remaining 32 rare materials collectively account for approximately 3.0\% of the dataset, providing diversity for edge cases and specialized physical behaviors.

This heterogeneous material distribution enables our model to learn a comprehensive physical prior spanning rigid bodies (metal, stone), deformable materials (rubber, foam), granular substances (sand, snow), and everyday materials (plastic, wood, fabric). The long-tail distribution also facilitates studying generalization to rare material types.
\textbf{Young's Modulus ($E$)}: Measures material stiffness, ranging from soft materials ($10^3$ Pa) to rigid materials ($4*10^{11}$ Pa). The dataset contains 10 distinct values spanning this range.

\textbf{Poisson's Ratio ($\nu$)}: Characterizes material compressibility, typically ranging from 0.01 to 0.49. The dataset includes 10 representative values covering common material behaviors.

\subsection{Source Datasets}
Our dataset aggregates models from the following four sources, each contributing unique characteristics:
\paragraph{OmniObject3D} A high-fidelity dataset featuring approximately 6,000 real-world scanned objects across 190 common categories (e.g., cups, chairs, animal models). It provides rich multi-modal data, including textured meshes with millimeter-level geometric accuracy and multi-view rendered images. For our purposes, we primarily leveraged its high-resolution rendered views (e.g., the 24-view set with associated camera parameters) to extract detailed appearance and geometric information.
\vspace{-0.3cm}
\paragraph{HSSD Dataset.}
The Habitat Synthetic Scenes Dataset (HSSD)~\cite{khanna2023hssd},  contains over 18,000 high-quality indoor scenes with photorealistic rendering and detailed semantic annotations. The dataset features diverse residential and commercial environments with realistic layouts and furnishings. 
\vspace{-0.3cm}
\paragraph{Amazon Berkeley Objects (ABO)} ABO offers a collection of approximately 8,000 high-quality, industry-standard 3D models covering 98 everyday object categories. The data includes textured CAD models (.obj/.glb), which we utilized to generate consistent multi-view renderings that align with our standardized format.
\vspace{-0.3cm}
\paragraph{Objaverse} A 10M+ dataset containing millions of 3D objects, Objaverse offers unparalleled diversity in object shape, category, and style. We selected a substantial subset from this collection to significantly broaden the scope and variety of our final dataset, as detailed in the following section.

\subsection{Data Processing}
\paragraph{Filter and Render}To ensure quality and consistency across the heterogeneous source datasets, we established a systematic data curation and processing pipeline. The Objaverse dataset, while extensive, is characterized by its considerable size and variable data quality. Consequently, to extract a high-quality subset, we employed a systematic curation strategy analogous to the one applied to gobjaverse. The screening procedure is outlined as follows: (1) A geometric similarity clustering algorithm was employed to identify and remove near-duplicate models. Any model exhibiting a similarity score of over 85\% with another was considered redundant and removed; (2) To filter out objects with non-standard or incomplete textures, we performed an analysis in the HSV (Hue, Saturation, Value) color space. Models where white pixels constituted more than 75\% of the surface texture were discarded, as this often indicates missing or placeholder textures. In the end, we filtered approximately about 20k data points in the Objaverse.For the other datasets, we used the full data without applying a filtering process.
For datasets that do not provide enough view rendering view, we use the rendering script provided by the corresponding dataset for enough view rendering. This procedure ensures that every object in our dataset is represented by a consistent set of views, capturing its complete geometric features for subsequent learning tasks.
% \vspace{-0.3cm}
% \paragraph{Physical Property Annotation} We employ a vision-language joint modeling approach for material classification. The eight orthogonal RGB images of each object are fed into a pre-trained Qwen3-VL~\cite{qwen3technicalreport} model. This model calculates the cross-modal similarity between the object's visual features and the textual descriptions of fourteen common materials (e.g., ``wood'', ``metal'', ``plasticine''). The final material label for each object is determined via a weighted voting mechanism based on these similarity scores. The prediction of mechanical parameters, specifically Young's modulus and the Poisson's ratio, is achieved through a similar methodology. Once a material is identified, its density value (in g/cm³) is assigned using a layered mapping framework that correlates the material label with standard physical constants.

\begin{figure*}[t]
\centering
\includegraphics[width=1.0\textwidth]{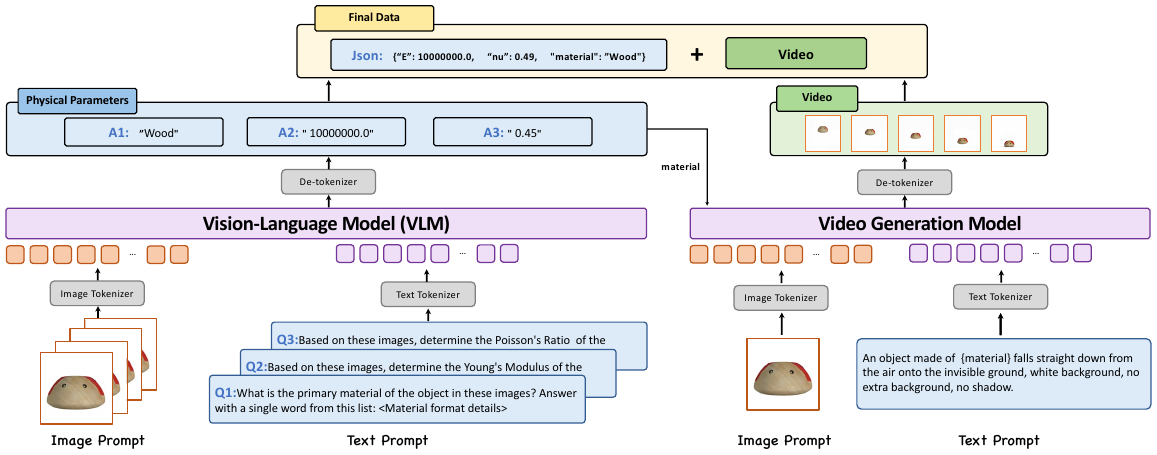}
\caption{Automated dataset construction pipeline. We predict physical properties using Qwen3-VL, and generate reference videos using FramePack. }
\label{fig:dataset_pipeline}
\end{figure*}

\section{Dataset Construction Pipeline.}
As shown in Figure~\ref{fig:dataset_pipeline}, we construct PhysAssets through 
an automated pipeline which predicting physical properties (material class, Young's modulus, Poisson's 
ratio) from 8 selected views using Qwen3-VL~\cite{qwen3technicalreport} and 
generating ground-truth reference videos using FramePack~\cite{zhang2025packing} 
conditioned on predicted properties. This pipeline enables scalable annotation of 
50,215 objects with physical properties and reference dynamics.

\subsection{Physical Property Annotation Pipeline}
\label{sec:property_annotation}

We develop a semi-automatic annotation pipeline leveraging multimodal large 
language models to predict three critical physical properties for each object: 
material class, Young's modulus ($E$), and Poisson's ratio ($\nu$). This 
approach enables scalable annotation of large-scale 3D datasets while maintaining 
consistency with real-world material physics.

\subsubsection{Visual Feature Extraction}

For each 3D object, we choose eight uniformly distributed views at fixed 
elevation. These multi-view RGB images provide comprehensive visual coverage of the object's geometry, texture, and appearance. The views are then fed into Qwen3-VL~\cite{qwen3technicalreport}, a state-of-the-art vision-language model pre-trained on diverse visual and textual data.

\paragraph{Material Classification}

Material classification is performed through vision-language alignment. We 
define a closed vocabulary of 45 primary materials commonly found in everyday objects: \textit{Wood, Metal, Plastic, Glass, Fabric, Leather, Ceramic, Stone, Rubber, Paper, Sand, Snow, Plasticine, Foam, etc.}. The model is queried with the following prompt:

\textbf{Material Classification Prompt:}

\textit{``What is the primary material of the object in these images? Answer 
with a single word from this list: Wood, Metal, Plastic, Glass, Fabric, Leather, 
Ceramic, Stone, Rubber, Paper, Sand, Snow, Plasticine, Foam, etc.''}

The model processes all eight views and outputs a material label based on 
cross-modal similarity between visual features and material descriptions. In 
cases of ambiguity, a weighted voting mechanism across views determines the 
final material class.

\paragraph{Young's Modulus Prediction}

Young's modulus ($E$) characterizes material stiffness—the resistance to 
elastic deformation under stress. We discretize the continuous range of 
Young's modulus values into 10 interpretable categories spanning from extremely soft materials (e.g., gel, foam) to ultra-stiff materials (e.g., diamond, 
tungsten). The model is prompted with:

\textit{``Based on these images, determine the Young's Modulus ($E$) of the object.}

\textit{What is Young's Modulus?}

\textit{Young's Modulus ($E$) measures a material's stiffness or resistance to 
elastic deformation. It indicates how much stress is needed to produce a given 
amount of strain (deformation).}

\textit{Select the most appropriate description:}

\begin{enumerate}[leftmargin=*,itemsep=2pt]

\item {\itshape extremely soft - Like gel or foam (e.g., jelly, soft foam) $\sim$ 1 KPa}
\item {\itshape very soft - Like rubber or sponge (e.g., rubber bands, foam mattress) $\sim$100 KPa}
\item {\itshape soft - Like soft plastics or leather (e.g., leather, soft PVC) $\sim$1 MPa}
\item {\itshape moderately soft - Like hard rubber (e.g., tire rubber) $\sim$10 MPa}
\item {\itshape moderate - Like nylon or cork (e.g., nylon, wood cork) $\sim$100 MPa}
\item {\itshape moderately stiff - Like hard plastics (e.g., ABS plastic, acrylic) $\sim$1 GPa}
\item {\itshape stiff - Like glass or ceramics (e.g., glass, porcelain) $\sim$10 GPa}
\item {\itshape very stiff - Like aluminum (e.g., aluminum, brass) $\sim$70 GPa}
\item {\itshape extremely stiff - Like steel (e.g., steel, iron) $\sim$200 GPa}
\item {\itshape ultra stiff - Like tungsten or diamond (e.g., tungsten, diamond) $\sim$400 GPa}

\end{enumerate}

\textit{Answer with ONLY ONE of these exact keywords.''}

The predicted categorical label is then mapped to a numerical value in Pascals 
(Pa) using the mapping defined in Table~\ref{tab:youngs_modulus_mapping}.

\paragraph{Poisson's Ratio Prediction}

Poisson's ratio ($\nu$) quantifies the ratio of lateral strain to axial strain 
when a material is deformed. We similarly discretize Poisson's ratio into 10 
categories representing different material behaviors, from auxetic materials 
(negative Poisson's ratio) to nearly incompressible materials (approaching 0.5). 
The prediction prompt is:

\textit{Poisson's Ratio Prediction Prompt:}

\textit{``Based on these images, determine the Poisson's Ratio ($\nu$) of the object.}

\textit{What is Poisson's Ratio?}

\textit{Poisson's Ratio ($\nu$) measures how much a material expands laterally 
when compressed axially, or contracts laterally when stretched. It describes 
the relationship between lateral strain and axial strain.}

\textit{Select the most appropriate description:}

\begin{enumerate}[leftmargin=*,itemsep=2pt]
\item {\itshape nearly incompressible - Almost no volume change (e.g., rubber) $\sim$0.50}
\item {\itshape high resistance - High lateral expansion (e.g., soft rubber) $\sim$0.45}
\item {\itshape moderately high - Moderately high deformation (e.g., gold, lead) $\sim$0.40}
\item {\itshape moderate high - Above average deformation (e.g., plastic, aluminum) $\sim$0.35}
\item {\itshape moderate - Typical for many metals (e.g., steel, iron) $\sim$0.30}
\item {\itshape moderate low - Below average deformation (e.g., glass) $\sim$0.25}
\item {\itshape low - Low lateral expansion (e.g., concrete, ceramics) $\sim$0.20}
\item {\itshape very low - Very low lateral expansion (e.g., cork) $\sim$0.15}
\item {\itshape extremely low - Minimal lateral deformation (e.g., foam) $\sim$0.10}
\item {\itshape auxetic - Negative Poisson's ratio materials $\sim$0.01}
\end{enumerate}

\textit{Answer with ONLY ONE of these exact keywords.''}

The categorical output is converted to a dimensionless numerical value using 
the mapping in Table~\ref{tab:poisson_ratio_mapping}.

\paragraph{Property Mapping Tables}

The categorical predictions from the vision-language model are mapped to 
numerical physical property values suitable for Material Point Method (MPM) 
simulation. Tables~\ref{tab:youngs_modulus_mapping} and~\ref{tab:poisson_ratio_mapping} 
present the complete mappings.

\begin{table}[h]
\centering
\caption{Young's Modulus categorical to numerical mapping. Values span 8 orders 
of magnitude, covering materials from soft gels to ultra-hard ceramics.}
\label{tab:youngs_modulus_mapping}
\small
\begin{tabular}{llr}
\toprule
\textbf{Category} & \textbf{Example Materials} & \textbf{Value (Pa)} \\
\midrule
extremely soft & Gel, foam, jelly & $1.0 \times 10^3$ \\
very soft & Rubber, sponge, silicone & $1.0 \times 10^5$ \\
soft & Leather, soft PVC, fabric & $1.0 \times 10^6$ \\
moderately soft & Hard rubber, tire rubber & $1.0 \times 10^7$ \\
moderate & Nylon, cork, paper & $1.0 \times 10^8$ \\
moderately stiff & Hard plastic (ABS, acrylic) & $1.0 \times 10^9$ \\
stiff & Glass, ceramic, porcelain & $1.0 \times 10^{10}$ \\
very stiff & Aluminum, brass, bronze & $7.0 \times 10^{10}$ \\
extremely stiff & Steel, iron, stainless steel & $2.0 \times 10^{11}$ \\
ultra stiff & Tungsten, diamond, carbide & $4.0 \times 10^{11}$ \\
\bottomrule
\end{tabular}
\end{table}

\begin{table}[h]
\centering
\caption{Poisson's Ratio categorical to numerical mapping. Values range from 
0.01 (auxetic materials) to 0.49 (nearly incompressible materials).}
\label{tab:poisson_ratio_mapping}
\small
\begin{tabular}{llr}
\toprule
\textbf{Category} & \textbf{Example Materials} & \textbf{Value} \\
\midrule
auxetic & Special engineered materials & 0.01 \\
extremely low & Foam materials & 0.10 \\
very low & Cork, engineered materials & 0.15 \\
low & Concrete, ceramics, brick & 0.20 \\
moderate low & Glass, cast iron & 0.25 \\
moderate & Steel, iron, brass, titanium & 0.30 \\
moderate high & Plastic, aluminum, copper & 0.35 \\
moderately high & Gold, lead, clay & 0.40 \\
high resistance & Soft rubber, flexible polymers & 0.45 \\
nearly incompressible & Rubber, elastomers & 0.49 \\
\bottomrule
\end{tabular}
\end{table}

\subsection{Video Generation and Preference Calculation} 
To facilitate the second stage of our training, which employs Direct Preference Optimization (DPO), we established a systematic pipeline for generating a dataset of preference tuples. This process is crucial for providing the high-quality, ranked data required to fine-tune our model on the nuances of physical dynamics. The pipeline consists of three main steps:
\subsubsection{Ground-Truth Video Generation}

We generate reference videos using FramePack~\cite{zhang2025packing}, 
guided by text prompts describing the physical scenario. After evaluating 
multiple prompt formulations (detailed below), we selected the following 
template for its optimal balance of simplicity and physical realism:

\begin{quote}
\textit{``An object made of \{material\} falls straight down from the air 
onto the invisible ground, white background, no extra background, no shadow.''}
\end{quote}

\subsubsection{Alternative Prompt Variants}

For reference and reproducibility, we document the alternative prompt variants 
explored during our experimentation. These prompts represent different trade-offs 
between prompt complexity, physical constraints, and generation control.

\paragraph{Prompt 2 (Detailed Physics Description):}
\begin{quote}\small
\textit{``A \{material\} toy centered on a plain pure white background. The 
\{material\} toy falls straight down vertically from the center of the frame 
to the bottom edge, obeying the laws of physics (gravity, acceleration). Show 
the entire descent: starting stationary at center, accelerating downwards, 
hitting the bottom edge with a subtle impact, and coming to a complete stop. 
The \{material\} toy remains rigid and inanimate throughout, showing no 
deformation or independent movement. Fixed, static camera view. No anthropomorphism, 
no unexpected motion, only the physics-based vertical fall and stop.''}
\end{quote}

\textit{Limitation:} Overly detailed constraints sometimes led to inconsistent 
generation or failure to satisfy all specified conditions.

\paragraph{Prompt 5 (Identity Preservation Focus):}
\begin{quote}\small
\textit{``Generate a short, high-fidelity video based on the provided object 
image, where the absolute highest priority is to strictly maintain the object's 
identity throughout the entire sequence. The scene features a seamless white 
background and a solid, invisible, horizontal white floor. The video begins 
with the object perfectly still in mid-air, then it is released to fall straight 
down vertically under gravity. Crucially, the object must maintain its initial 
orientation during the fall, without any tumbling, spinning, or rotation. The 
object is made of \{material\}, and its impact and subsequent behavior must 
realistically simulate the physical properties of this material.''}
\end{quote}

\textit{Limitation:} While improving identity preservation, this prompt 
occasionally resulted in unrealistic motion due to strict orientation constraints.

\paragraph{Prompt 6 (Photorealism Emphasis):}
\begin{quote}\small
\textit{``Generate a short, photorealistic video based on the provided input 
image, simulating the object falling and impacting the ground. Throughout the 
entire video, the object must retain its original visual identity—its shape, 
texture, and color. The fall itself must be completely inanimate and passive; 
the object must descend in a pure vertical drop without any rotation, spinning, 
or tumbling. Upon impact with the flat white ground, the object's physical 
reaction must precisely mimic the properties of \{material\}. The entire event 
takes place in a seamless, infinite white studio environment.''}
\end{quote}

\textit{Limitation:} We found that excessively long prompts with detailed constraints often compromise generation quality, leading to inconsistent or unnatural motion.

The selected prompt consistently produced the most physically plausible and visually 
coherent videos across diverse materials and object geometries.

\subsubsection{Candidate Video Generation} Leveraging the model pre-trained in Stage 1, we generate a set of plausible, yet varied, simulation outcomes. Specifically, we sample three distinct sets of physical properties (e.g., Young's modulus, Poisson's ratio) from the learned probability distribution associated with the object. Each of these property sets is then used to run a new simulation, producing three unique candidate videos that represent different potential physical behaviors.
\subsubsection{Preference Labeling via Trajectory Alignment}
\label{sec:preference_labeling}

To create preference pairs for DPO training, we develop an automatic labeling 
pipeline that compares simulated dynamics against reference videos through 
three-stage trajectory alignment. We employ 
SAM-2~\cite{ravi2024sam} for object segmentation and CoTracker-3~\cite{karaev24cotracker3} 
for dense trajectory extraction across both ground-truth and simulated sequences.
\vspace{-0.3cm}
\paragraph{Spatial Alignment.}
Due to different camera viewpoints and object scales between reference and simulated videos, direct trajectory comparison is infeasible. We address this 
through bounding box normalization: for each video, we compute the object's 
bounding box from its segmentation mask as $\mathcal{B} = (x_{\min}, y_{\min}, 
x_{\max}, y_{\max})$. Point trajectories from the ground-truth video are first 
normalized to $[0,1]$ coordinates relative to its bounding box:
\begin{equation}
\mathbf{p}_{\text{norm}} = \left( \frac{x - x_{\min}}{x_{\max} - x_{\min}}, 
\frac{y - y_{\min}}{y_{\max} - y_{\min}} \right)
\end{equation}
These normalized coordinates are then mapped to the simulated video's coordinate 
frame using its bounding box parameters. This spatial alignment ensures 
correspondence between trajectories regardless of viewpoint or scale differences.
\vspace{-0.3cm}
\paragraph{Landing Frame Alignment.}
Physical simulations may exhibit different temporal dynamics (e.g., falling 
speeds) even with similar physical properties. To enable fair comparison, we 
align sequences based on a key physical event: the object's landing moment. 
Specifically, we identify the landing frame as the temporal turning point where 
the object's vertical motion reverses. For each video, we track the point with 
maximum $y$-coordinate in the first frame (typically the object's bottom) and 
monitor its trajectory. The landing frame $f^*$ is detected when the vertical 
velocity changes sign:
\begin{equation}
f^* = \arg\min_{t} \{ t \mid y_t \leq y_{t-1}, t > 0 \}
\end{equation}
where $y_t$ represents the tracked point's $y$-coordinate at frame $t$. This 
frame marks the transition from falling to resting/bouncing phases.
\vspace{-0.3cm}
\paragraph{Temporal Alignment.}
Using the detected landing frames $(f^*_{\text{GT}}, f^*_{\text{sim}})$ as 
temporal anchors, we align the post-landing phases of both sequences. We 
determine the comparable duration as $T = \min(T_{\text{GT}} - f^*_{\text{GT}}, 
T_{\text{sim}} - f^*_{\text{sim}})$, where $T_{\text{GT}}$ and $T_{\text{sim}}$ 
are the total frame counts. Additionally, we compute a spatial offset 
$(\Delta x, \Delta y)$ between the landing positions in both videos and apply 
this correction to the simulated trajectories:
\begin{equation}
\mathbf{p}_{\text{sim}}^{\text{aligned}} = \mathbf{p}_{\text{sim}} + (\Delta x, \Delta y)
\end{equation}
This ensures that both sequences are aligned not only temporally but also 
spatially at the critical landing event.
\vspace{-0.3cm}
\paragraph{Similarity Metric.}
After three-stage alignment, we compute the trajectory dissimilarity as:
\begin{equation}
d(\mathcal{V}_{\text{sim}}, \mathcal{V}_{\text{GT}}) = \frac{1}{NT} 
\sum_{n=1}^{N} \sum_{t=1}^{T} \|\mathbf{p}_{n,t}^{\text{GT}} - 
\mathbf{p}_{n,t}^{\text{sim}}\|_2
\end{equation}
where $N$ is the number of tracked points and $T$ is the aligned sequence length. 
Lower dissimilarity indicates better physical plausibility. For each scene, we 
rank $K$ candidate simulations by this metric and select the best as the 
``winner'' and worst as the ``loser'' for DPO training.

\subsubsection{Additional Data Sources}
It is also worth noting that the PhysX~\cite{cao2025physx} dataset was released concurrently with our research, offering 3D objects annotated with physical properties ,which is also suitable for our dataset process. Given the timing constraints, its integration was not feasible for the present study. Nevertheless, we acknowledge its significance and view it as a promising avenue for extending our work in the future.
\begin{figure}[ht]
\centering
\includegraphics[width=0.4\textwidth]{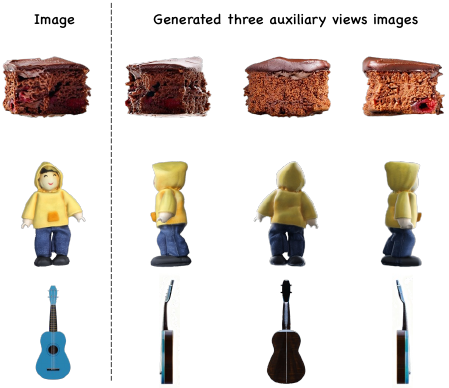} % Reduce the figure size so that it is slightly narrower than the column.
\caption{\textbf{Multi-view generation using MVAdapter.} Given a single frontal view image as input (left), MVAdapter~\cite{huang2025mv} generates three auxiliary views: rear, left, and right (right three panels). These 
synthesized views, together with the input frontal view, provide comprehensive angular coverage for our 3D Gaussian reconstruction and physics prediction pipeline. The generated views maintain consistent geometry and appearance while capturing different perspectives of the object.}
\label{fig6}
\end{figure}
\section{Additional Results}
\begin{figure*}[ht]
\centering
\includegraphics[width=0.75\textwidth]{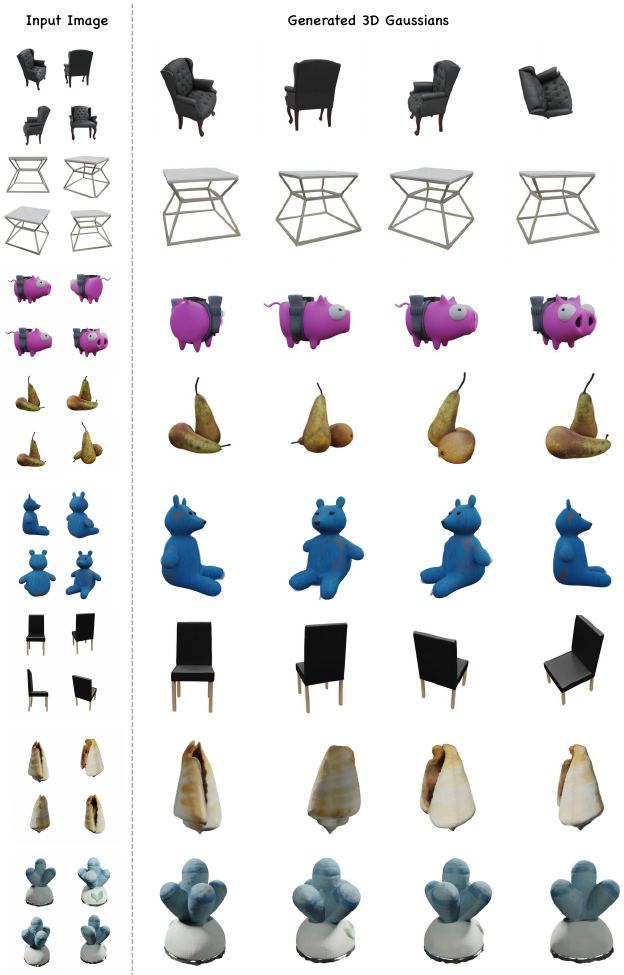} % Reduce the figure size so that it is slightly narrower than the column.
\caption{Qualitative results for Multi-View Stereo. Our method generates Gaussian splatting with remarkable visual quality on various challenging images.}
\label{fig5}
\end{figure*}

\begin{figure*}[ht]
\centering
\includegraphics[width=1.0\textwidth]{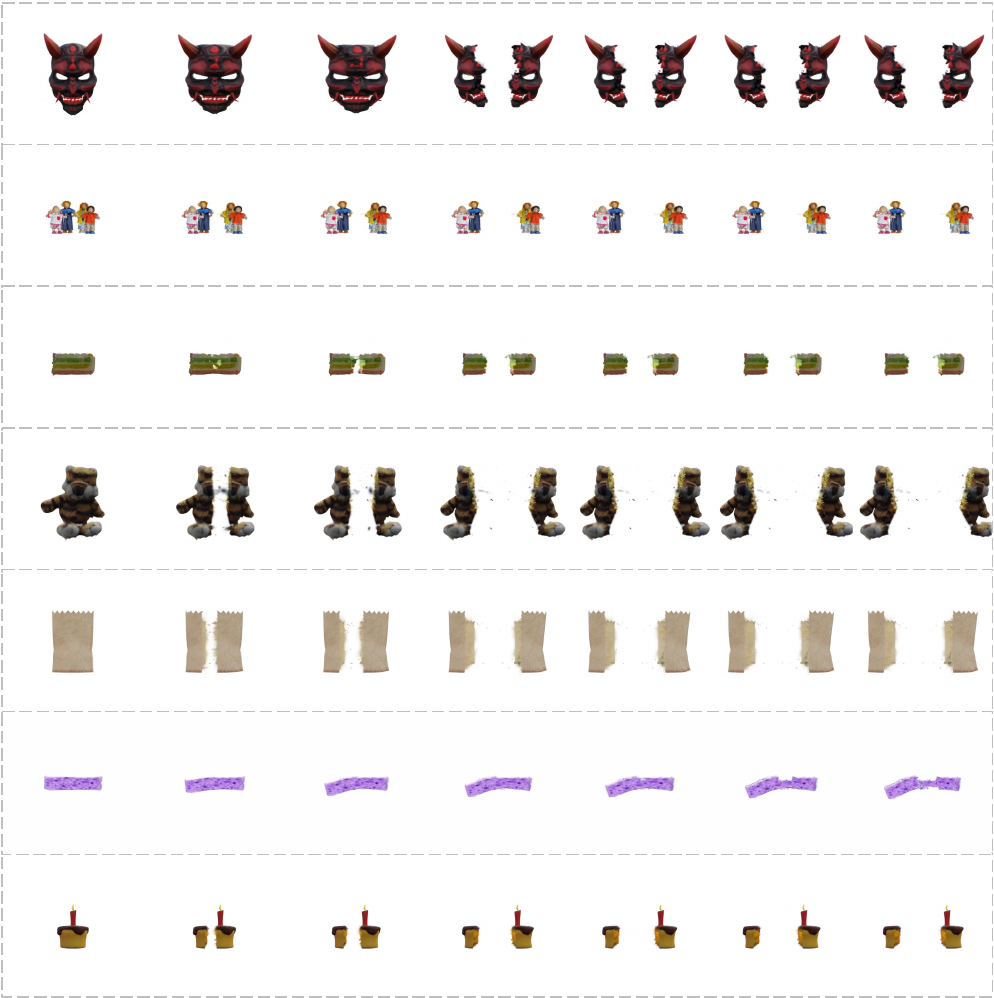} % Reduce the figure size so that it is slightly narrower than the column.
\caption{Qualitative results for stretching scenarios. Our method correctly captures the distinct responses of different materials under tensile forces.}
\label{fig13}
\end{figure*}

\begin{figure*}[ht]
\centering
\includegraphics[width=0.9\textwidth]{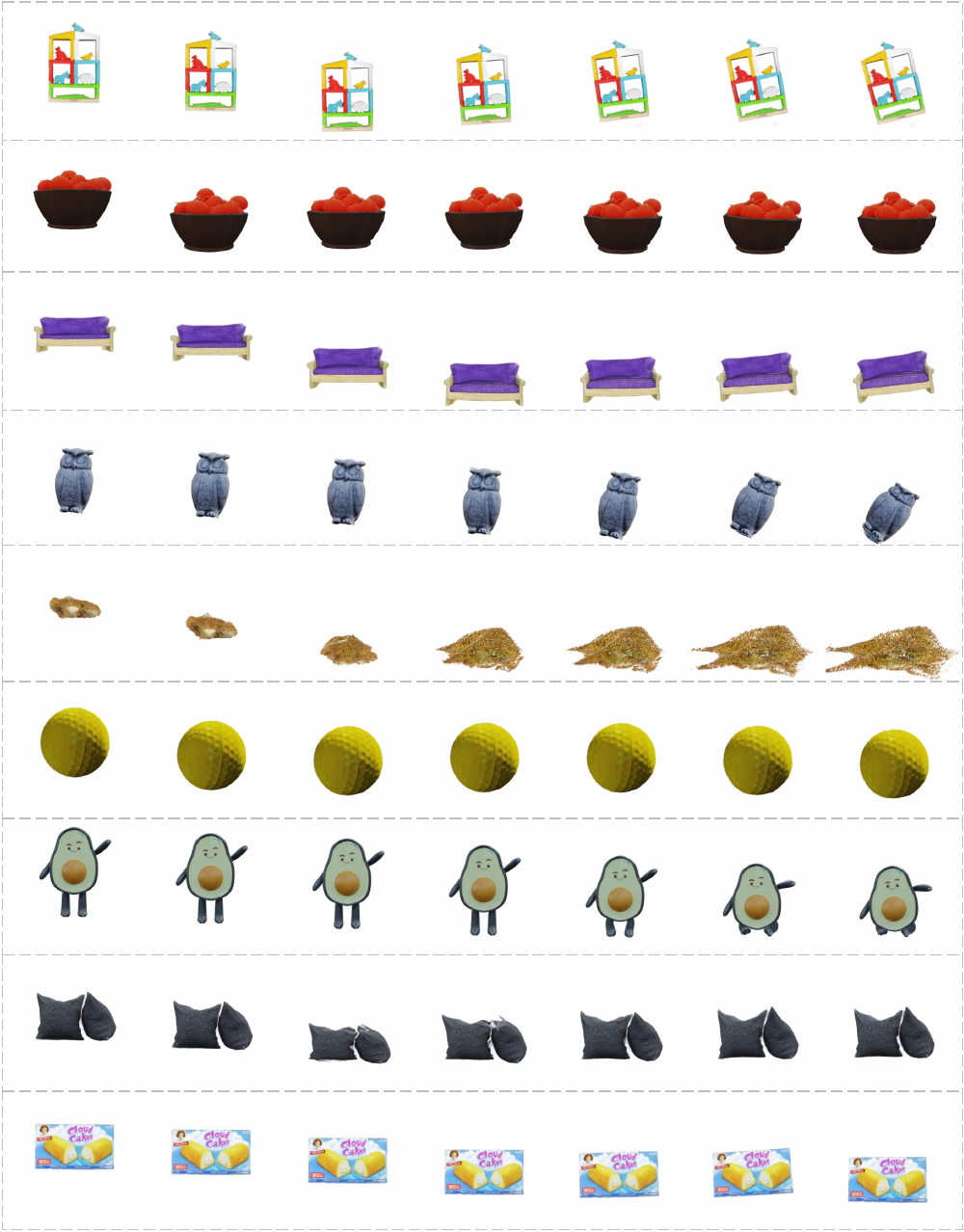} % Reduce the figure size so that it is slightly narrower than the column.
\caption{Qualitative results for dropping scenarios. Our model accurately predicts the physical properties of different materials, leading to plausible deformation and final states upon impact with the ground.}
\label{fig14}
\end{figure*}

\begin{figure*}[ht]
\centering
\includegraphics[width=1.0\textwidth]{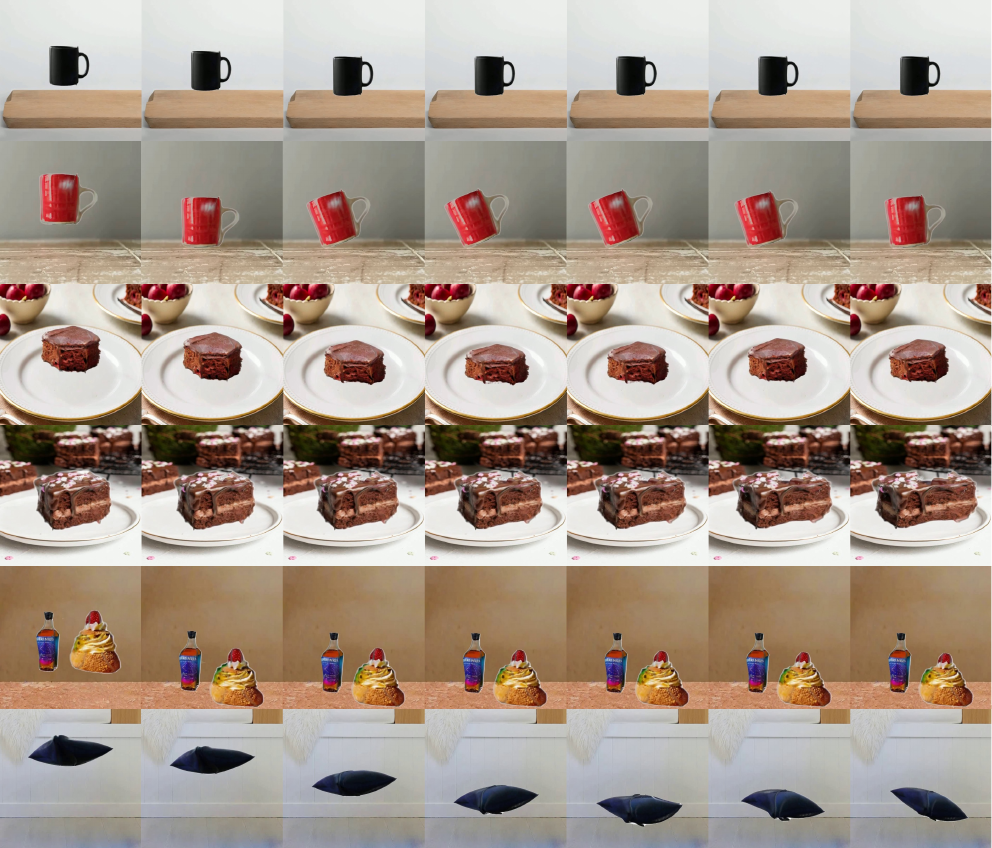} % Reduce the figure size so that it is slightly narrower than the column.
\caption{Demonstration of our model's robustness in in-the-wild scenes. } 
\label{fig15}
\end{figure*}

\begin{figure*}[ht]
\centering
\includegraphics[width=1.0\textwidth]{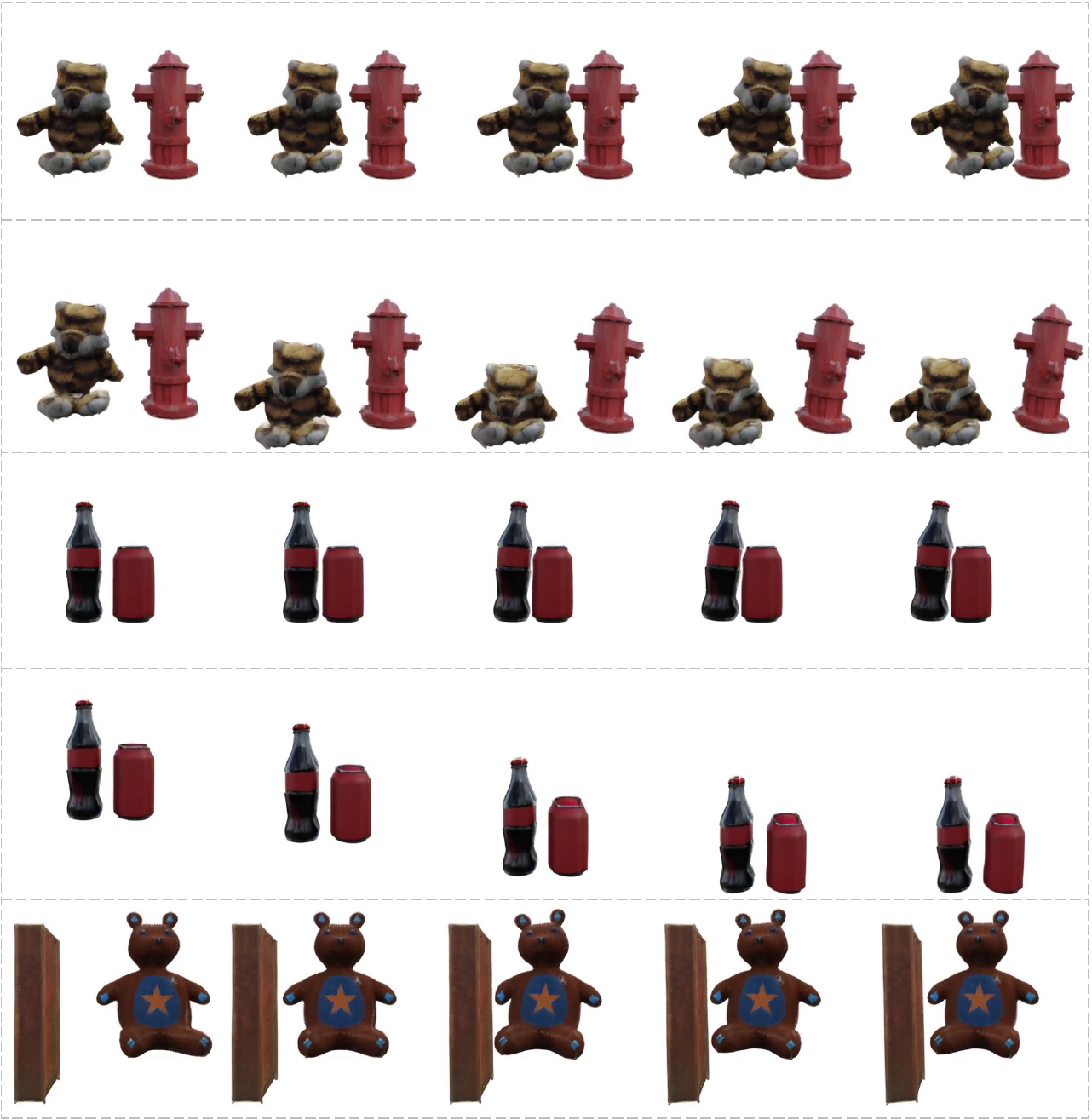} % Reduce the figure size so that it is slightly narrower than the column.
\caption{Qualitative results for multi-object interaction scenarios. Our approach can handle more complex scenes involving simultaneous collisions and interactions, generating physically consistent outcomes for all objects.}
\label{fig16}
\end{figure*}

\begin{figure*}[ht]
\centering
\includegraphics[width=1.0\textwidth]{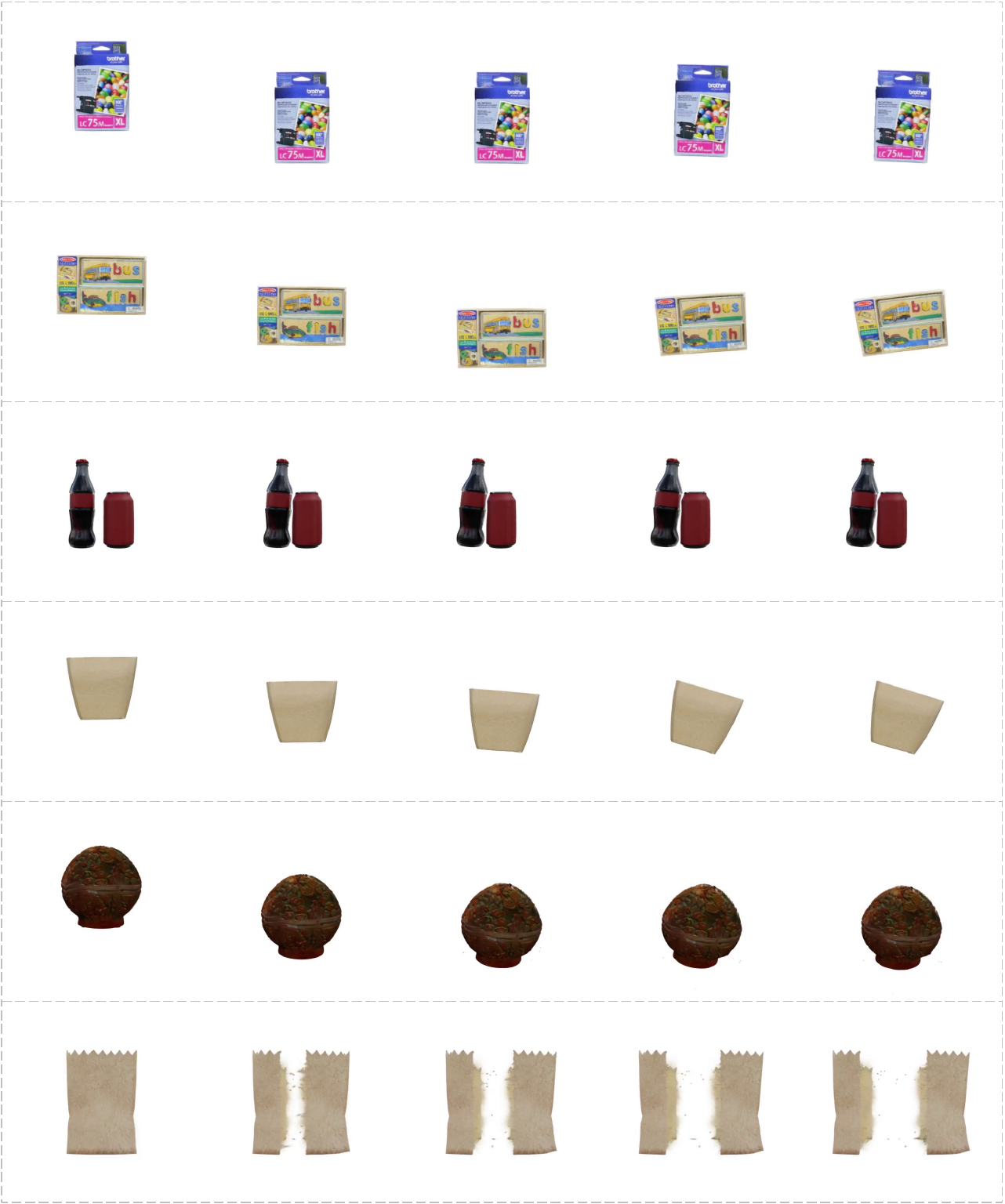} % Reduce the figure size so that it is slightly narrower than the column.
\caption{Other results.}
\label{fig10}
\end{figure*}

\begin{figure*}[ht]
\centering
\includegraphics[width=1.0\textwidth]{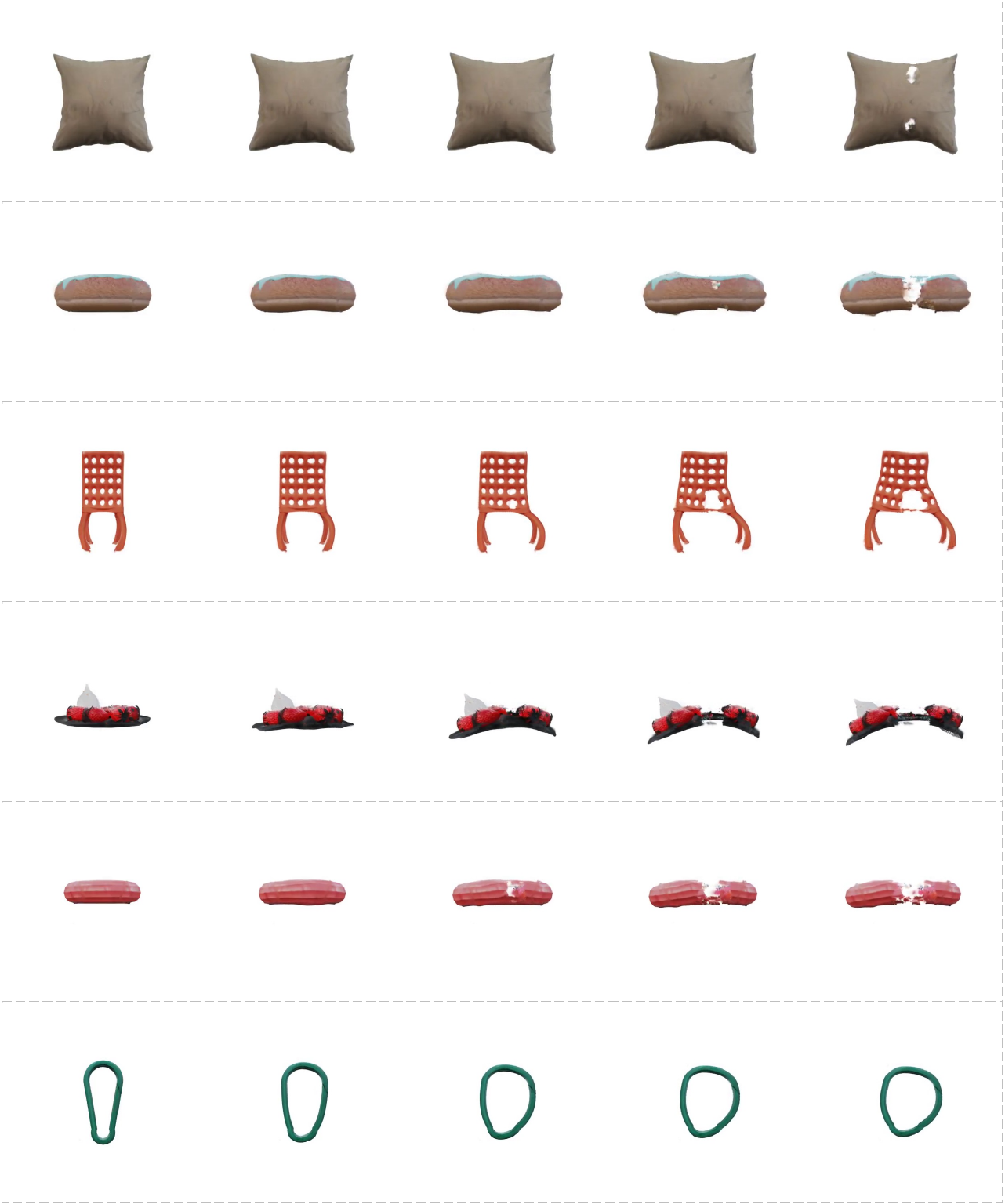} % Reduce the figure size so that it is slightly narrower than the column.
\caption{Other results.}
\label{fig11}
\end{figure*}

\begin{figure*}[ht]
\centering
\includegraphics[width=1.0\textwidth]{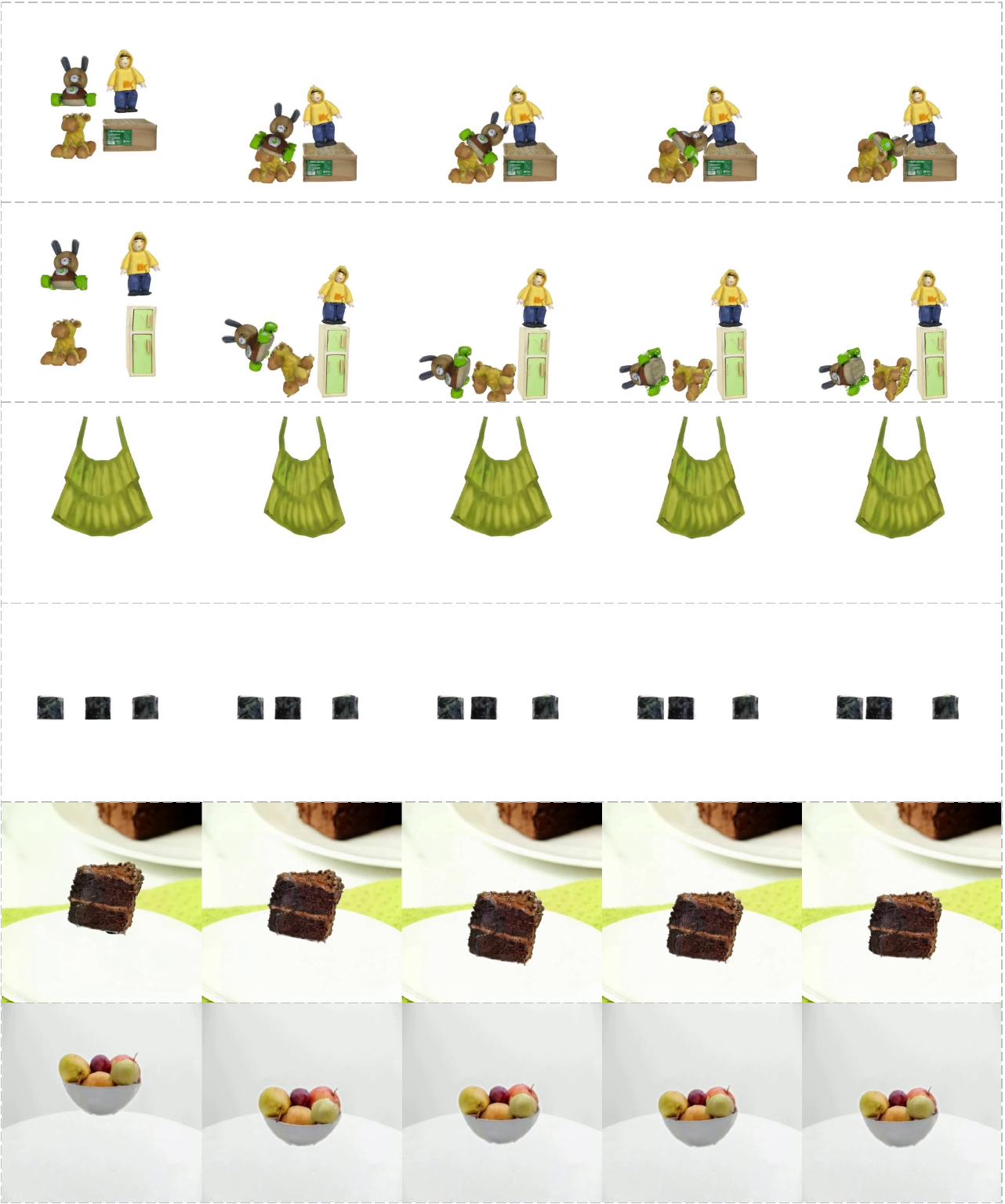} % Reduce the figure size so that it is slightly narrower than the column.
\caption{Other results.}
\label{fig12}
\end{figure*}

To fully demonstrate the versatility and effectiveness of our approach, we present an extended suite of supplementary experiments with comprehensive qualitative insights. Specifically, Figure~\ref{fig5} provides detailed visualizations of the Multi-View Stereo 
(MVS) module, showcasing its exceptional ability to accurately reconstruct 
3D geometry from multi-view inputs. Given four randomly selected input views, 
the model generates novel viewpoints, and we visualize four representative 
views sampled from eight randomly selected output views as examples. Complementarily, Figure~\ref {fig6} offers visualizations of the MVAdapter component, clearly revealing how it effectively bridges domain gaps and enhances feature alignment across diverse input modalities. Beyond the core component validations, Figure~\ref {fig13} and Figure~\ref {fig14} exhibit the model’s performance on fundamental stretching and dropping scenarios, respectively. We further push the envelope to validate its effectiveness under more challenging configurations: Figure~\ref {fig15} illustrates strong robustness in cluttered/complex background scenes, while Figure~\ref {fig16} highlights its superior capability in handling intricate multi-object interactions and other results in Figure~\ref {fig10}~\ref {fig11}~\ref {fig12}.

\section{Limitations and Future Work}

While PhysGM demonstrates significant advances in fast, physically-grounded 3D synthesis, it is important to acknowledge its current limitations, which also highlight promising directions for future research.
\vspace{-0.3cm}
\paragraph{Data Dependency and Generalization.}
Our model's performance is inherently tied to the scope and diversity of the PhysAssets dataset. While large, the dataset primarily consists of rigid objects. Consequently, the model may not generalize well to out-of-distribution categories, such as highly deformable or articulated objects. Future work could focus on expanding the dataset and exploring domain adaptation techniques to handle a wider variety of object types.
\vspace{-0.3cm}
\paragraph{Simplified Physics Representation.}
PhysGM currently predicts a single, ``lumped'' vector of physical properties (e.g., one mass, one friction coefficient) for the entire object. This assumes uniform material composition, which is not true for many real-world objects (e.g., a hammer with a metal head and wooden handle). A significant next step would be to extend our framework to predict spatially varying material properties, enabling more complex and realistic simulations.

% {
%     \small
%     \bibliographystyle{ieeenat_fullname}
%     \bibliography{main}
% }
\clearpage

% \clearpage

% \section{Rationale}
% \label{sec:rationale}
% % 
% Having the supplementary compiled together with the main paper means that:
% % 
% \begin{itemize}
% \item The supplementary can back-reference sections of the main paper, for example, we can refer to \cref{sec:intro};
% \item The main paper can forward reference sub-sections within the supplementary explicitly (e.g. referring to a particular experiment); 
% \item When submitted to arXiv, the supplementary will already included at the end of the paper.
% \end{itemize}
% 
% To split the supplementary pages from the main paper, you can use \href{https://support.apple.com/en-ca/guide/preview/prvw11793/mac#:~:text=Delete%20a%20page%20from%20a,or%20choose%20Edit%20%3E%20Delete).}{Preview (on macOS)}, \href{https://www.adobe.com/acrobat/how-to/delete-pages-from-pdf.html#:~:text=Choose%20%E2%80%9CTools%E2%80%9D%20%3E%20%E2%80%9COrganize,or%20pages%20from%20the%20file.}{Adobe Acrobat} (on all OSs), as well as \href{https://superuser.com/questions/517986/is-it-possible-to-delete-some-pages-of-a-pdf-document}{command line tools}.

\clearpage
\paragraph{Acknowledgements.} This work was supported by the NSFC under Grants U2441242. This work is also supported by the National Natural Science Foundation of China (U23B2013 and 62276176)

{
    \small
    \bibliographystyle{ieeenat_fullname}
    \bibliography{main}
}

\end{document}